\documentclass[10pt,twocolumn,letterpaper]{article}

\usepackage{cvpr}              

\usepackage{graphicx}
\usepackage{amsmath}
\usepackage{amssymb}
\usepackage{booktabs}
\usepackage{float}

\usepackage{siunitx}

\usepackage{bm}
\usepackage[dvipsnames]{xcolor}

\usepackage{algorithm}
\usepackage[noend]{algpseudocode}
\usepackage{multirow}
\usepackage{booktabs}

\usepackage{xcolor}
\definecolor{Cyan}{cmyk}{1,0,0,0}

\newcommand{\review}[1]{{\color{black}#1}}

\usepackage[pagebackref,breaklinks,colorlinks]{hyperref}

\usepackage[capitalize]{cleveref}
\crefname{section}{Sec.}{Secs.}
\Crefname{section}{Section}{Sections}
\Crefname{table}{Table}{Tables}
\crefname{table}{Tab.}{Tabs.}

\def\cvprPaperID{133} 
\def\confName{CVPR}
\def\confYear{2022}

\begin{document}

\title{UPST-NeRF: Universal Photorealistic Style Transfer of Neural Radiance Fields for 3D Scene}

\author{
{Yaosen Chen}\textsuperscript{1},
%
{Qi Yuan}\textsuperscript{1},
{Zhiqiang Li}\textsuperscript{1},
{Yuegen Liu}\textsuperscript{1,3}, 
{Wei Wang}\textsuperscript{*1,2},
{Chaoping Xie}\textsuperscript{1,2},\\
{Xuming Wen}\textsuperscript{1,2},
{and Qien Yu}\textsuperscript{4}
\\
\\
\textsuperscript{1}Media Intelligence Laboratory, ChengDu Sobey Digital Technology Co., Ltd \\
\textsuperscript{2}Peng Cheng Laboratory \\
\textsuperscript{3}Southwest Jiaotong University \\
\textsuperscript{4}Sichuan University \\
{\tt\small \{chenyaosen, yuanqi, lizhiqiang, liuyuegen, wangwei, xiechaoping, wenxuming\}@sobey.com}
\\
{\tt\small yuqien@scu.edu.cn}
}

\twocolumn[{%
\renewcommand\twocolumn[1][]{#1}%

\maketitle

\vspace{-9mm}
\begin{center}
    \centering
    \captionsetup{type=figure}
    \includegraphics[width=0.95\linewidth]{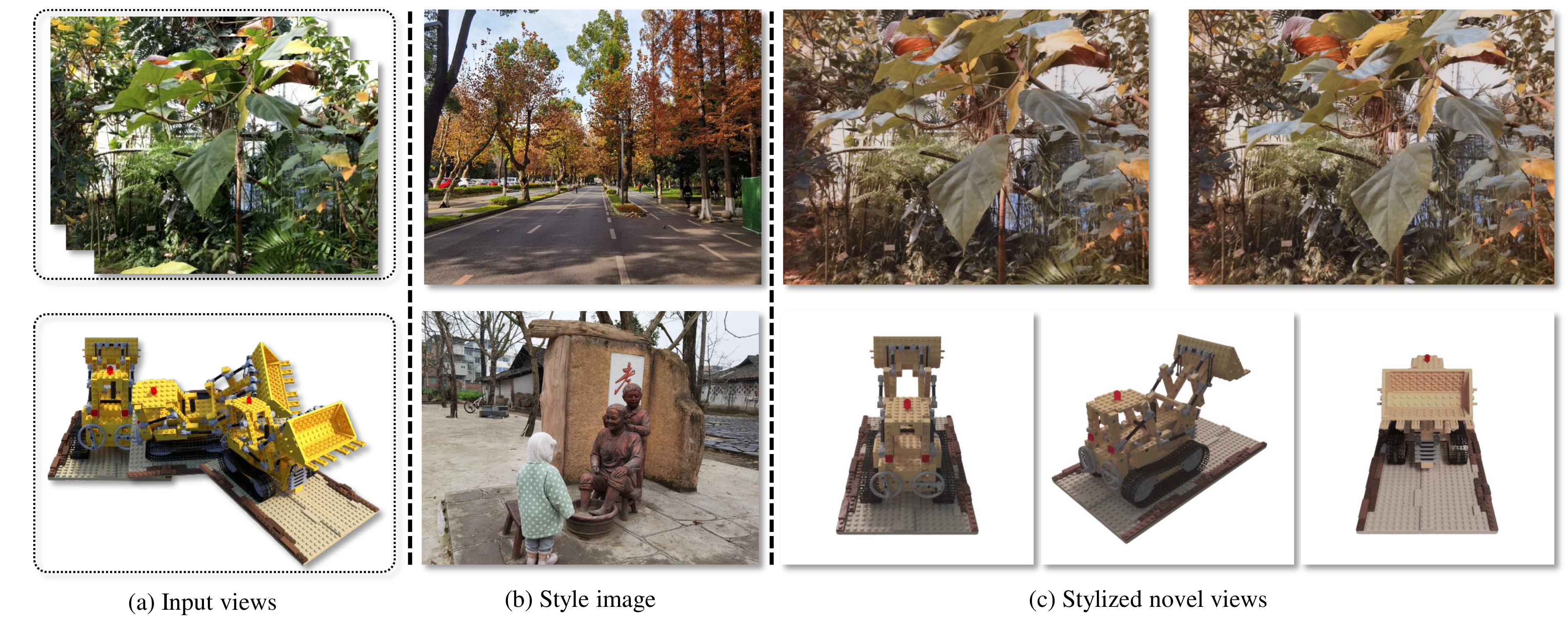}
    \vspace{-4mm}
    \captionof{figure}{\textbf{Transferring photorealistic style with a style image in the 3D scene}. Multi-view images of a given set of 3D scenes (a) and a style image (b), our model is capable of rendering photorealistic stylized novel views  (c) with a consistent appearance at various view angles in 3D space. }
\label{fig:teaser}
\end{center}
}]

\if TT\insert\footins{\noindent\footnotesize{
*Corresponding Author is Wei Wang (wangwei@sobey.com).}}\fi

\begin{abstract}

3D scenes photorealistic stylization aims to generate photorealistic images from arbitrary novel views according to a given style image while ensuring consistency when rendering from different viewpoints. Some existing stylization methods with neural radiance fields can effectively predict stylized scenes by combining the features of the style image with multi-view images to train 3D scenes. However, these methods generate novel view images that contain objectionable artifacts. Besides, they cannot achieve universal photorealistic stylization for a 3D scene. Therefore, a styling image must retrain a 3D scene representation network based on a neural radiation field. We propose a novel 3D scene photorealistic style transfer framework to address these issues. It can realize photorealistic 3D scene style transfer with a 2D style image. We first pre-trained a 2D photorealistic style transfer network, which can meet the photorealistic style transfer between any given content image and style image. Then, we use voxel features to optimize a 3D scene and get the geometric representation of the scene. Finally, we jointly optimize a hyper network to realize the scene photorealistic style transfer of arbitrary style images. In the transfer stage, we use a pre-trained 2D photorealistic network to constrain the photorealistic style of different views and different style images in the 3D scene. The experimental results show that our method not only realizes the 3D photorealistic style transfer of arbitrary style images but also outperforms the existing methods in terms of visual quality and consistency.  Project
page:\href{https://semchan.github.io/UPST_NeRF/}{https://semchan.github.io/UPST\_NeRF/}.
\end{abstract}

\section{Introduction}
\label{sec:intro}

In recent years, the 3D implicit representation method based on the neural radiation field~\cite{mildenhall2020nerf,zhang2020nerf++} has made great progress because of its excellent performance in scene realism. By controlling the appearance in these scenes, style transfer can reduce the time of artistic creation and the need for professional knowledge. Many excellent works achieve this goal through texture generation~\cite{xiang2021neutex,kanazawa2018learning,gao2020tmnet} and semantic view synthesis~\cite{habtegebrial2020generative,huang2020semantic}. Some recent work~\cite{huang2021learning,hollein2022stylemesh,huang2022stylizednerf,chiang2022stylizing,nguyen2022snerf,zhang2022arf,fan2022unified} can transfer artistic features from a single 2D image to a complete real 3D scene, thereby changing the style in the real scene.

Most of these methods focus on how to solve the consistency problem of stylized scenes. LSVN~\cite{huang2021learning} proposed a point cloud-based method for consistent 3D scene stylization.
StyleMesh~\cite{hollein2022stylemesh} optimized an explicit texture for the reconstructed mesh of a scene and stylized it jointly from all available input images. StylizedNeRF~\cite{huang2022stylizednerf} proposed a mutual learning framework for 3D scene stylization, which combines a 2D image stylization network and NeRF to fuse the stylization ability of a 2D stylization network with the 3D consistency of NeRF. To solve the blurry results and inconsistent appearance, Stylizing-3D-Scene~\cite{chiang2022stylizing} utilized a hyper network to transfer the style information into the scene representation. To eliminate the jittering artifacts due to the lack of cross-view consistency, SNeRF~\cite{nguyen2022snerf} investigated 3D scene stylization that provides a strong inductive bias for consistent novel view synthesis. ARF~\cite{zhang2022arf} proposed to stylize the more robust radiance field representation and produce consistent stylized novel views of high visual quality. Besides, INS~\cite{fan2022unified} conducted a pilot study on various implicit functions, including 2D coordinate-based representation, neural radiance field, and signed distance function. These methods only realize the transfer of artistic style, if photorealistic images are used as style images to transfer the style of 3D scenes, the stylized scenes will contain objectionable artifacts.

This paper aims to stylize a photorealistic 3D scene following a given set of style examples, and our method allows generating stylized images of the scene from arbitrary novel views while ensuring rendered images from different viewpoints are consistent. To ensure consistency, we formulate the problem as stylizing a NeRF~\cite{mildenhall2020nerf} with a given set of style images. Some examples of our stylization method are presented in Fig.~\ref{fig:teaser}.

Neural radiance fields (NeRF)~\cite{mildenhall2020nerf} use multi-layer perceptron (MLP)  to implicitly learn the mapping from the queried 3D point with its colors and densities to reconstruct a volumetric scene representation from a set of images. This method dramatically improves the quality of scene rendering, but it requires a long training time and inefficient novel view rendering. To reconstruct the scene quickly, we are inspired by DirectVoxelGO ~\cite{sun2022direct} and use the voxel grid to directly optimize the geometric appearance of the scene in our geometric training stage. It contains two voxel grids: one is the density voxel grid, which is used to predict the occupancy probability; the other is the feature voxel grid, which is followed by a shallow MLP(RGBNet) for color predicting.

Since the implicit continuous volumetric representation is built on the millions of voxel grid parameters, it is unclear which parameters control the style information of the 3D scene. To overcome this issue, one possible solution is combining existing image/video stylization approaches with novel view rendering techniques~\cite{chiang2022stylizing} which firstly novel view images and then perform image stylization. Inspired by Stylizing-3D-Scene~\cite{chiang2022stylizing}, we use a HyperNet and HyperLinear to handle the ambiguities of the 2D stylized learnable latent codes as conditioned inputs. Unlike Stylizing-3D-Scene, we use StyleEncoder with VGG~\cite{simonyan2014very} to extract style features from the style images. Then we use the style features as the input of HyperNet to update the weights of HyperLiner. Finally, we use HyperLiner to change the information of RGBNet to achieve the style of updating the scene. To present more realistically, instead of directly using Adaptive Instance Normalization (AdaIN)~\cite{huang2017arbitrary} to constrain the style of the novel view, we trained a 2D photorealistic style transfer network to process truth value RGB under different views to obtain the target, which use to constrain the predicted color value.

In a nutshell, our main contributions are as follows:
\begin{itemize}
	\item We propose a novel universal photorealistic style transfer of neural radiance fields for photorealistic stylizing 3D scenes with given style images.
    \item We propose a hyper network to control the features of photorealistic style images as the latent codes of scene stylization and use the 2D method to realize the geometric consistency constraint of the neural radiation field.
    \item To realize the scene's photorealistic style transfer, we designed an efficient 2D style transfer network to process the 2D photorealistic style images under different novel views to constrain the scene style. 
\end{itemize}

\section{Related Work}
\label{sec:relate}

\noindent \textbf{Novel View Synthesis.} Novel view synthesis aims to generate the images at arbitrary viewpoints from a set of source images.
Some studies use a single image or a pair of stereo images as input and use methods such as Multi Plane Image (MPI)~\cite{srinivasan2019pushing,tucker2020single,wizadwongsa2021nex,zhou2018stereo,mildenhall2019local}, light field techniques~\cite{gortler1996lumigraph,davis2012unstructured,levoy1996light}, point cloud~\cite{niklaus20193d,wiles2020synsin} to represent the scene for synthesizing novel view near the input viewpoint.
However, these methods cannot generate a novel view image from an arbitrary viewpoint. To generate images of novel view image from arbitrary viewpoints , these methods need more images as input to reconstruct the scene.
Some works build 3D scenes by combining geometric representation with color~\cite{waechter2014let,seitz1999photorealistic}, texture~\cite{debevec1996modeling}, light field~\cite{buehler2001unstructured,wood2000surface} or neural rendering~\cite{flynn2019deepview,lombardi2019neural,sitzmann2019deepvoxels,niklaus20193d,wiles2020synsin,meshry2019neural,aliev2020neural}. This 3D implicit representation method based on neural radiance fields (NeRF)~\cite{mildenhall2020nerf} greatly improves the quality of novel view generation. Subsequently, some work extended NeRF to octree structure~\cite{liu2020neural}, unbounded scenes~\cite{zhang2020nerf++}, reflectance decomposition~\cite{boss2021nerd} and uncontrolled real-world images~\cite{martin2021nerf}.
However, NeRF and its variants require a training time from hours to days for a single scene, making it infeasible for many application scenarios. In view of this, DirectVoxelGO ~\cite{sun2022direct} uses gradient-descent to optimize voxel grids directly predict the grid values and can rapidly train from scratch in less than 15 minutes with a single GPU.

\begin{figure*}[htbp]
	\centering
	\includegraphics[width=0.95\linewidth]{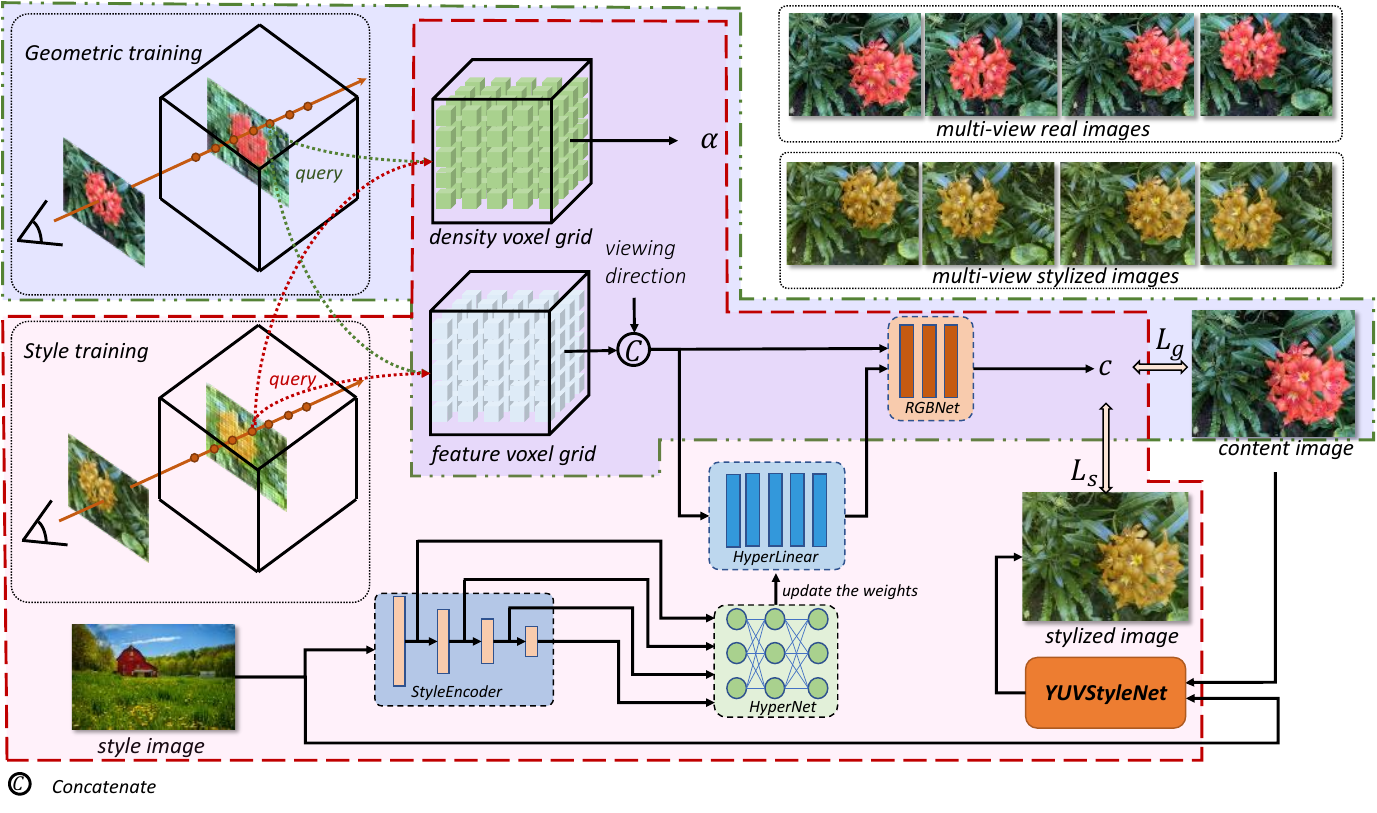}
	\caption{\textbf{Overview of Universal Photorealistic Style Transfer of Neural Radiance Fields.} In our framework, the training in photorealistic style transfer in 3D scenes divides into two stages. The first stage is geometric training for a single scene. We use the density voxel grid and feature voxel grid to represent the scene directly, and the density voxel grid is used to output density; the feature voxel grid with a shallow MLP of RGBNet use to predict the color. The second stage is style training. The parameters of the density voxel grid and feature voxel grid will be frozen, and we use reference style image's features to be the input of the hyper network, which can control the RGBNet's input. Thus, we jointly optimize the hyper network to realize the scene photorealistic style transfer with arbitrary style images.}
	\label{fig:framework}
	\vspace{-5mm}
\end{figure*}
\noindent \textbf{Image and video style transfer.}
There are two important categories of style transfer tasks: artistic style transfer and photorealistic style transfer. Using Gram matrix~\cite{gatys2015neural} can transfer the style information from the reference image to the content image. It is widely used in the task of artistic style transfer. For faster stylization, Avatar ~\cite{sheng2018avatar}, and AdaIN~\cite {huang2017arbitrary} leverage feed-forward neural networks. DPST~\cite{luan2017deep} proposed a deep photorealistic style transfer method by constraining the transformation to be locally affine in colorspace. To improve the efficiency, PhotoWCT ~\cite{li2018closed}, WCT$^2$~\cite{yoo2019photorealistic} have been proposed. Xia et al.~\cite{xia2020joint}propose an end-to-end model for photorealistic style transfer that is both fast and inherently generates photorealistic results. Qiao et al. ~\cite{qiao2021efficient} proposed Style-Corpus Constrained Learning (SCCL) to relieve the unrealistic artifacts and heavy computational cost issues.

To ensure the consistency between adjacent frames, and make the stylized video not flicker, optical flow or temporal constraint-based methods ~\cite{gao2020fast,chen2017coherent} are applied to video stylization. MCCNet~\cite{deng:2020:arbitrary} can be trained to fuse the exemplar style features and input content features for efficient style transfer and achieves coherent results. Wang et al. proposed jointly considering the intrinsic properties of stylization and temporal consistency for video style transfer. However, these 2D-based methods lack spatial consistency constraints and 3D scene perception, so they cannot maintain long-term consistency in 3D scene style transfer.

\noindent \textbf{3D sence style transfer.}
Through texture generation~\cite{xiang2021neutex,kanazawa2018learning,gao2020tmnet} and semantic view synthesis~\cite{habtegebrial2020generative,huang2020semantic} can editing the appearance in 3D scenes. Using an image as a reference and changing the style of the scene has also become a hot topic of recent research for 3D sense style transfer. Spatial consistency becomes one of the main problems to be solved in 3D scene stylization. For example, LSVN~\cite{huang2021learning} proposed a point cloud-based method for consistent 3D scene stylization, and Stylizing-3D-Scene~\cite{chiang2022stylizing} utilized a hyper network to transfer the style into the scene to solve the blurry results and inconsistent appearance. StyleMesh~\cite{hollein2022stylemesh} stylized the 3D scene jointly from all available input images and optimized an explicit texture for scene reconstruction. StylizedNeRF~\cite{huang2022stylizednerf} utilize the stylization ability of 2D stylization network and neural radiation field for 3D scene stylization, and ARF~\cite{zhang2022arf} proposed to stylize the more robust radiance field representation. SNeRF~\cite{nguyen2022snerf} investigated 3D scene stylization, providing a strong inductive bias for consistent novel view synthesis. INS~\cite{fan2022unified} studied unifying the style transfer for 2D coordinate-based representation, neural radiance field, and signed distance function. These methods can achieve the artistic style transfer in 3D scenes, but it isn't easy to realize the photorealistic style transfer. When these methods are applied to photorealistic style transfer, they will lead to artifacts when rendering the synthesis of a novel view.

\begin{figure*}[htbp]
	\centering
	\includegraphics[width=0.85\linewidth]{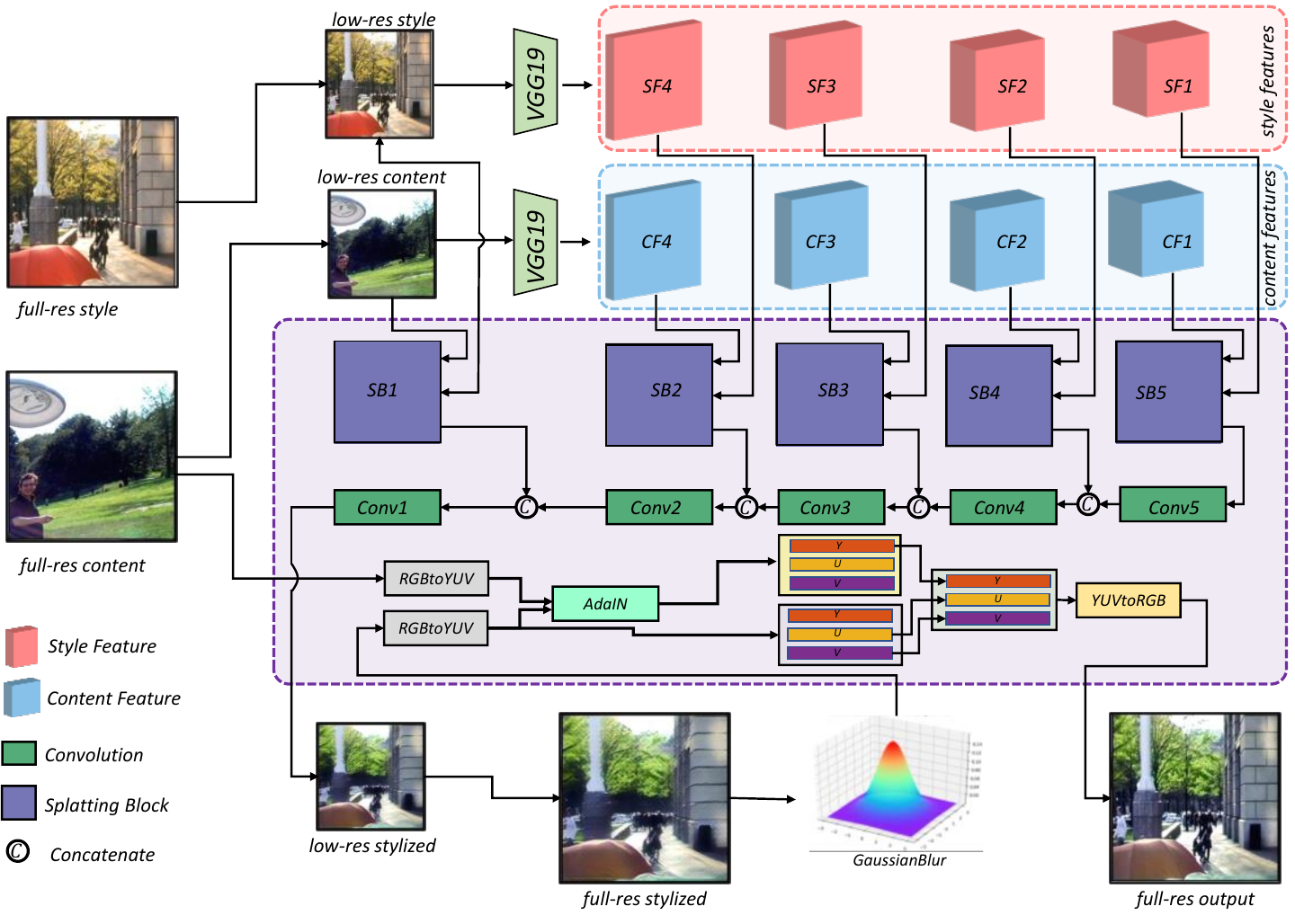}
	\caption{\textbf{The architecture of YUVStyleNet.} We designed a framework for 2D photorealistic style transfer, which supports the input of a full resolution style image and a full resolution content image, and realizes the photorealistic transfer of styles from the style image to the content image. In this framework, we transform the image into YUV channels. The final fusion uses the generated stylized UV channel, and the Y channel fusion after the stylized image is fused with the original content image to get the final photorealistic stylized image.}
	\label{fig:YUVStyleNet}
	\vspace{-4mm}
\end{figure*}

\section{Preliminaries}

NeRF~\cite{mildenhall2020nerf} employ multiplayer perceptron (MLP) networks to model a scene as a continuous volumetric field of opacity and radiance. One MLP, indicated as $\operatorname{MLP}^{(pos)}$, for density predicting  and the other MLP, indicated as $\operatorname{MLP}^{(rgb)}$, for radiance color  predicting:
\begin{subequations}
	\label{eq:nerf_mlp}
	\begin{align}
		(\sigma, \bm{e}) &= \operatorname{MLP}^{\mathrm{(pos)}}(\operatorname{PE}(\bm{x}))~,\\
		\bm{c} &= \operatorname{MLP}^{\mathrm{(rgb)}}(\bm{e}, \operatorname{PE}(\bm{d}))~,
	\end{align}
\end{subequations}
where $\bm{x}\in\mathbb{R}^3$ is the 3D position, $\bm{d}\in\mathbb{R}^2$ is the viewing direction, $\sigma\in\mathbb{R}^+$ is the corresponding density, $\bm{c}\in\mathbb{R}^3$ is the view-dependent color emission, $\bm{e}\in\mathbb{R}^{D_{e}}$ is an intermediate embedding tensor with dimension ${D_{e}}$ and $\operatorname{PE}$ is positional encoding.

The ray $\bm{r}$ from the camera center through the pixel for rendering the color of a pixel $C(\bm{r})$:
\begin{subequations} \label{eq:volume_rendering}
	\begin{align}
		{C}(\bm{r}) &= \left( \sum_{i=1}^{K} T_i \alpha_i \bm{c}_i \right) + T_{\scriptscriptstyle K+1} \bm{c}_{\mathrm{bg}}  ~, \\
		\alpha_i &= \operatorname{alpha}(\sigma_i, \delta_i) = 1 - \exp(-\sigma_i \delta_i) ~, \label{eq:density_2_alpha} \\
		T_i &= \prod_{j=1}^{i-1} (1 - \alpha_j) ~, \label{eq:acc_trans}
	\end{align}
\end{subequations}
where $K$ is the number of sampling points on $\bm{r}$ between the pre-defined near and far planes; $\alpha_i$ is the probability of termination at the point $i$; $T_i$ is the accumulated transmittance from the near plane to point $i$; $\delta_i$ is the distance to the adjacent sampled point, and $\bm{c}_{bg}$ is a pre-defined background color.

In the training stage, NeRF optimizes the model by minimizing the Mean Square Error (MSE) between the pixel color $C(\bm{r})$ of the image in the training set and the rendered pixel color $C_{gt}(\bm{r})$. 
\begin{equation} \label{eq:photo_loss}
	\mathcal{L}_{\mathrm{mse}} = \frac{1}{|\mathcal{R}|} \sum_{r\in\mathcal{R}} \left\|{C}(\bm{r}) - C_{gt}(\bm{r})\right\|_2^2 ~ ,
\end{equation}
where $\mathcal{R}$ is the set of rays in a sampled mini-batch.

\section{Our Approach}

The overview of the universal photorealistic style transfer of neural radiance fields has been shown in Fig.~\ref{fig:framework}. Through several images of a given scene, our goal is to generate a photorealistic styled image of an arbitrary viewpoint in the scene according to the reference style image while maintaining geometric consistency. In our framework, for a single scene, we achieve the training through two stages: geometric training and style training. Then, in the rendering processing, we can synthesize novel viewpoints with photorealistic style transfer according to the style of an arbitrary reference image.

\subsection{Scene Geometric Reconstruction} \label{sec:SceneGeometricReconstruction}

Similar to DirectVoxelGO ~\cite{sun2022direct}, we adopt voxel grid to represent the 3D scene. Such an scene representation is efficient to query for any 3D positions via interpolation:
\begin{equation} \label{eq:interp}
	\operatorname{interp}(\bm{x}, \bm{V}): \left(\mathbb{R}^3, \mathbb{R}^{C \times N_x \times N_y \times N_z}\right) \to \mathbb{R}^C ~,
\end{equation}
where $\bm{x}$ is the queried 3D point, $\bm{V}$ is the voxel grid, $C$ is the dimension of the modality, and $N_x\cdot N_y \cdot N_z$ is the total number of voxels.
Trilinear interpolation is applied if not specified otherwise.

\begin{equation} \label{eq:pre_in_post}
		\alpha =\operatorname{alpha}(\operatorname{softplus}(\operatorname{interp}(\bm{x},\bm{V}^{\text{(density)}} )))
\end{equation}
where $\operatorname{alpha}$ (\cref{eq:density_2_alpha}) functions sequentially for volume rendering, $\operatorname{softplus}$ is the activation function and $\bm{V}^{\text{(density)}}  \in \mathbb{R}^{1 \times N_x \times N_y \times N_z }$ is the density voxel grid.

For view-dependent color emission predicting can be expressed as:
\begin{equation} \label{eq:pre_in_post1}
    \bm{c} = \operatorname{MLP}^{\text{(rgb)}}\left(\operatorname{interp}(\bm{x}, \bm{V}^{\text{(feat)}}), \bm{x}, \bm{d}\right)
\end{equation}

where $\bm{c} \in \mathbb{R}^3$ is the view-dependent color emission, $\bm{V}^{\text{(feat)}}  \in \mathbb{R}^{D \times N_x \times N_y \times N_z }$ is the feature voxel grid, $\operatorname{D}$ is a hyperparameter for feature-space dimension. By default, we set D equal to $128$. The MLP is shown in Fig. \ref{fig:framework} as RGBNet.

We use the photometric loss in Eq.\ref{eq:photo_loss}. Similar to DirectVoxelGO ~\cite{sun2022direct}, we incorporate per-point rgb loss and background entropy loss and the modification loss as below:
 \begin{equation} \label{eq:pt_rgb}
     \mathcal{L}_{\mathrm{pt\_rgb}} = \frac{1}{|\mathcal{R}|} \sum_{r\in\mathcal{R}} \sum_{i=1}^{K} \left( T_i \alpha_i \left\|\bm{c}_i - C(\bm{r})\right\|_2^2~ \right).
 \end{equation}
 The background entropy loss regularizes the rendered background probability, $T_{\scriptscriptstyle K+1}$ in \cref{eq:volume_rendering}, to concentrate on either foreground or background:
 \begin{equation}\label{eq:bg}
     \mathcal{L}_{\mathrm{bg}} = -T_{\scriptscriptstyle K+1}\log(T_{\scriptscriptstyle K+1}) - (1-T_{\scriptscriptstyle K+1})\log(1-T_{\scriptscriptstyle K+1})~.
 \end{equation}
 Finally, the overall training objective of the geometric training stage is
 \begin{equation} \label{eq:lg}
     {L_{g}} =
     {\lambda}_{\mathrm{photo}}^{\text{(c)}}\cdot \mathcal{L}_{\mathrm{photo}} +
     {\lambda}_{\mathrm{pt\_rgb}}^{\text{(c)}}\cdot \mathcal{L}_{\mathrm{pt\_rgb}} +
     {\lambda}_{\mathrm{bg}}^{\text{(c)}}\cdot \mathcal{L}_{\mathrm{bg}}~,
 \end{equation}
 where ${\lambda}_{\mathrm{photo}}^{\text{(c)}}, {\lambda}_{\mathrm{pt\_rgb}}^{\text{(c)}}, {\lambda}_{\mathrm{bg}}^{\text{(c)}}$ are hyper parameters of the loss weights.
\vspace{4mm}
\subsection{YUVStyleNet for 2D photorealistic stylization} \label{sec:2d_style}
2D photorealistic stylization is the task of transferring the style of a reference image onto a content target, which makes a photorealistic result that is plausibly taken with a camera. In our work, we designed a 2D photorealistic style transfer network for the photorealistic style transfer of images from a novel view of the scene to get a photorealistic style transfer network in the style training stage. We name the 2D photorealistic style transfer network YUVStyleNet, and its detailed framework shows in Fig. ~\ref{fig:YUVStyleNet}.

To reduce the GPU memory and improve the processing speed, the input full-resolution style image ${I_f^s}$ and the content image ${I_f^c}$ will be downsampled to the corresponding low-resolution images ${I_l^s}$ and ${I_l^c}$. In our experiment, the size of low-resolution image is $512$ by default. Then, we use a pre-trained VGG model to extract style features $\{F_j^s\}|1 \leq j \leq  4$ and content features $\{F_j^c\}|1 \leq j \leq  4$ at different scales respectively. The style features and content features of the corresponding scale, as well as the corresponding low-resolution style image and content image, are used as feature pairs as the input of the splatting block module to obtain the output $\{F_i^{sb}\}|1 \leq i \leq  5$  under the corresponding scale. 

We first extract the input $s$ feature and $c$ feature through a convolution and then use adaptive instance normalization (AdaIN)~\cite{huang2017arbitrary} to fuse the features in the splatting block module. Splatting block output features are concatenated with low-scale features, respectively, and then through convolution operation, the low resolution stylized image ${I_l^{sed}}$ is finally obtained. By upsampling, we get a stylized image ${I_f^{sed}}$ with the same scale as the input ${I_f^{c}}$. To make the color transfer smoother in space, we use a Gaussian filter to process ${I_f^{sed}}$ and get ${I_f^{sg}}$. We convert the original content image and ${I_f^{sg}}$ to YUV domain as ${I_f^{cyuv}}$ and ${I_f^{sgyuv}}$ , and then get ${I_f^{scyuv}}$ through AdaIN. To make the generated photorealistic stylized image consistent with the original image in brightness, we extract Y channels from ${I_f^{scyuv}}$ and UV channels from ${I_f^{sgyuv}}$ and to get a new style image ${I_f^{sedyuv}}$. Then the final ${I_f^{sedrgb}}$ is obtained by converting to RGB space.

We refer to AdaIN~\cite{huang2017arbitrary} to define our style loss ${L_s}$ and content loss ${L_c}$. In addition, to obtain a more photorealistic effect, We compared Peak Signal to Noise Ratio (PSNR) and Structural Similarity (SSIM) as constraints between the generated style image ${I_f^{sedrgb}}$  and the original content image ${I_f^{c}}$, we add PSNR loss ${L_{psnr}}$ and SSIM loss ${L_{ssim}}$:
 \begin{equation} \label{eq:yuv_loss}
	{L_{yuvs}} = {\lambda}_{c}\cdot{L_c} +
	{\lambda}_{s}\cdot{L_s} +
	{\lambda}_{psnr}\cdot{L_{psnr}} +
	{\lambda}_{ssim}\cdot{L_{ssim}} 
\end{equation}
 
We randomly select content image and style image in MS-COCO~\cite{2014Microsoft} to train YUVStyleNet, and finally optimize to get a better effect of photorealistic style transfer.

\subsection{Style Learning in 3D Scene} \label{sec:3d_style_learning}
To reconstruct and change the style of the scene by using arbitrary style images as input, we designed a hyper network (HyperNet) and a hyper linear network (HyperLinear) to control the input features of RGBNet when rendering the scene. As shown in Fig.\ref{fig:framework}, in the style training stage, the feature queried by the feature voxel grid under the corresponding view is spliced with the view direction feature as the input of HyperLinear network, and the output of HyperLinear will be directly used as the input of RGBNet to control the generation of color. The style image is extracted through the pre-trained feature extraction network, VGGNet, and then used as the input of HyperNet. The output of HyperNet is used to control the weight of HyperLinear, to modify the scene's color through style features.
\begin{figure}[htbp]
	\centering
	\includegraphics[width=0.95\linewidth]{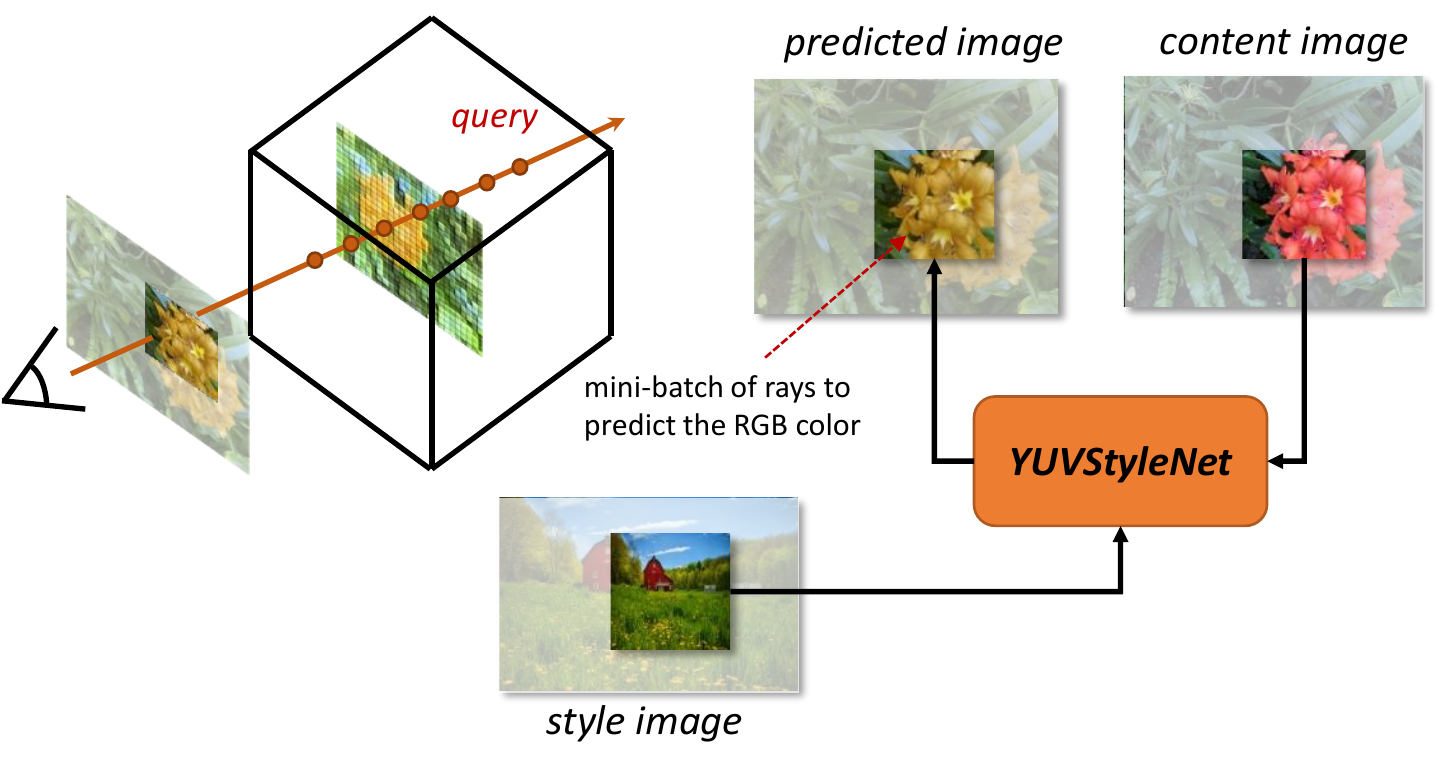}
	\caption{\review{\textbf{RGB color predicting with mini-batch of rays in style training}
		}
	}
	\label{fig:mini-batch-predicting}
\end{figure}

In the stage of 3D style training, we constrain the training process of style transfer by optimizing Eq. \ref{eq:pt_rgb}, \ref{eq:bg} and \ref{eq:lg}. The difference from that introduced in Section ~\ref{sec:SceneGeometricReconstruction} is that we have changed $C(\bm{r})$ with YUVStyleNet. We get the corresponding content image through a mini-batch of rays,demonstrate in Fig.~\ref{fig:mini-batch-predicting}, and the style image randomly collected from MS-COCO~\cite{2014Microsoft} is used as the input of YUVStyleNet, and the predicted image is used as  $C_{s}(\bm{r})$.
Therefore, Eq. \ref{eq:pt_rgb} can be adjusted to:
 \begin{equation} \label{eq:style_pt_rgb}
	\mathcal{L}_{\mathrm{s\_pt\_rgb}} = \frac{1}{|\mathcal{R}|} \sum_{r\in\mathcal{R}} \sum_{i=1}^{K} \left( T_i \alpha_i \left\|\bm{c}_i - C_{s}(\bm{r})\right\|_2^2~ \right).
\end{equation}
Thus, the loss constraint of scene style transfer is:
 \begin{equation} \label{eq:style_lg}
	{L_{s}} =
	{\lambda}_{\mathrm{photo}}^{\text{(c)}}\cdot \mathcal{L}_{\mathrm{photo}} +
	{\lambda}_{\mathrm{s\_pt\_rgb}}^{\text{(c)}}\cdot \mathcal{L}_{\mathrm{s\_pt\_rgb}} +
	{\lambda}_{\mathrm{bg}}^{\text{(c)}}\cdot \mathcal{L}_{\mathrm{bg}}.
\end{equation}
where ${\lambda}_{\mathrm{photo}}^{\text{(c)}}, {\lambda}_{\mathrm{s\_pt\_rgb}}^{\text{(c)}}, {\lambda}_{\mathrm{bg}}^{\text{(c)}}$ are hyper parameters of the loss weights.
\vspace{4mm}

In the style training stage, we froze the density voxel grid and feature voxel grid, which is optimized in the process of geometry training. At the same time, we also froze the parameters of YUVStyleNet, and the parameters of StyleEncoder, which uses VGG as the decoder for feature extraction.
 
\begin{figure*}[htbp]
	\centering
	\includegraphics[width=1.\linewidth]{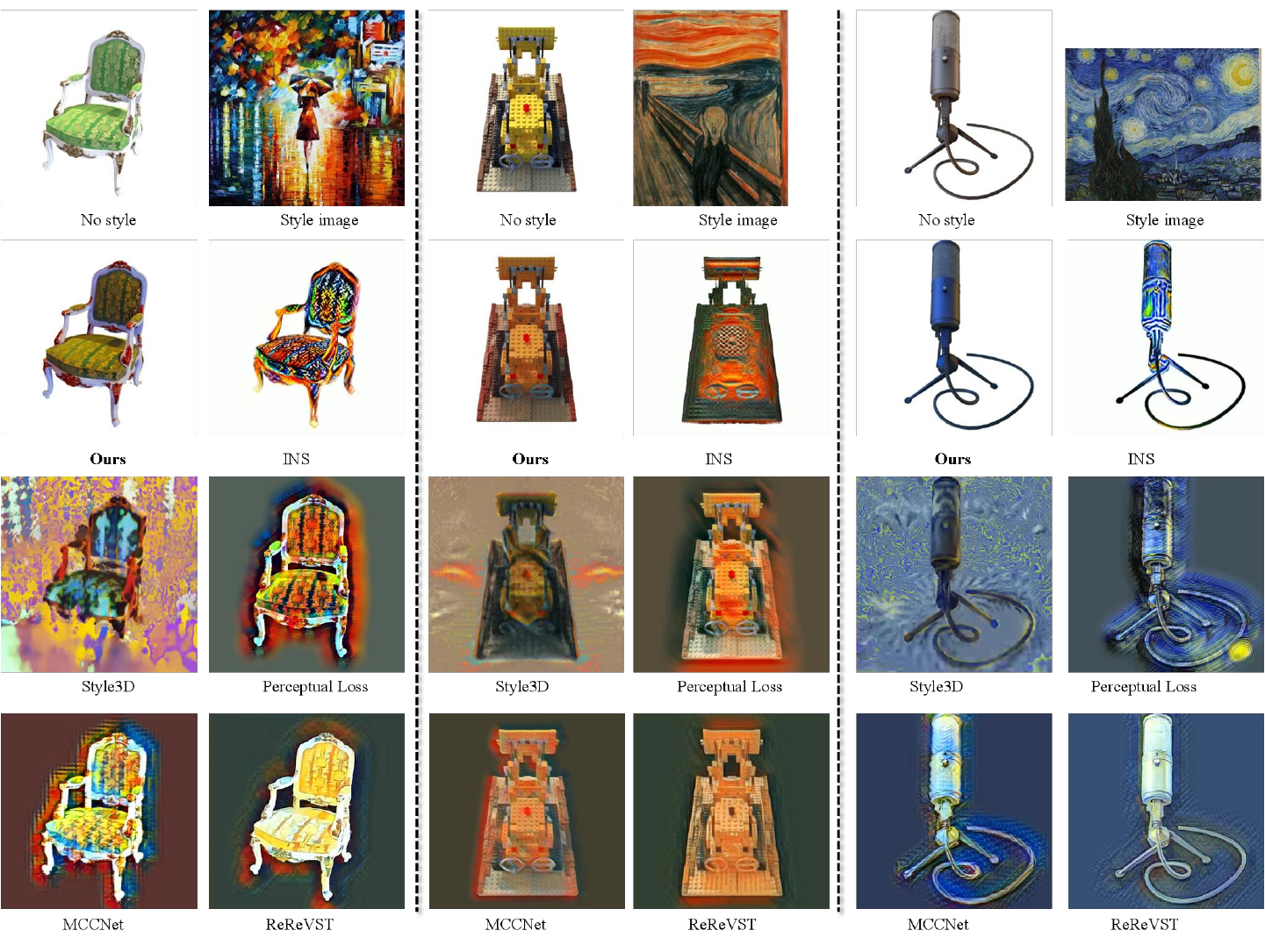}
	\caption{\textbf{Qualitative comparisons with artistic style images}. We compare the stylized results of 3 scenes on NeRF-Synthetic dataset. Our method stylizes scenes with more precise geometry and competitive stylization quality.}
	\label{fig:st_c_a}
	\vspace{-4mm}
\end{figure*}
\begin{figure*}[htbp]
	\centering
	\includegraphics[width=1.\linewidth]{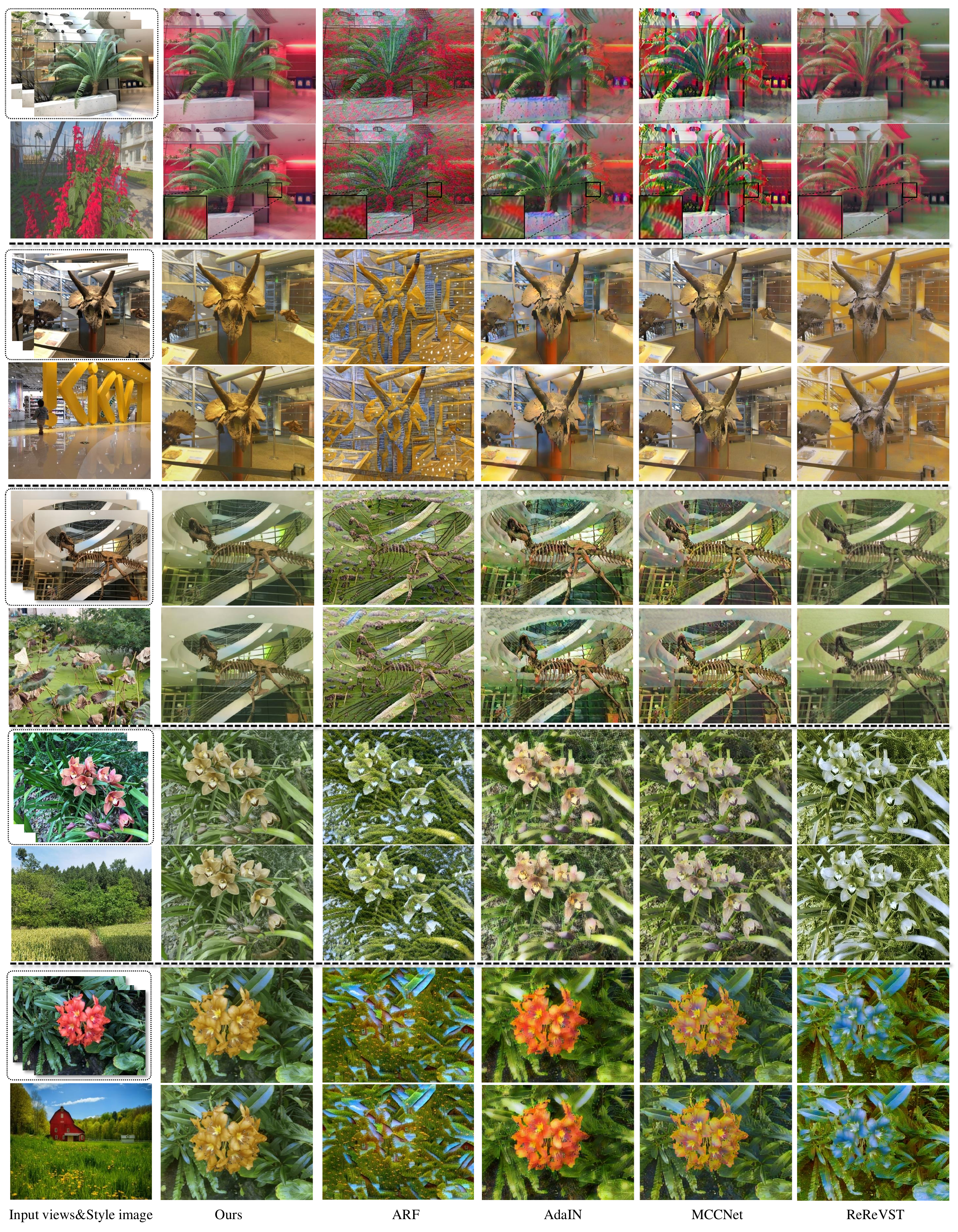}
	\caption{\textbf{Qualitative comparisons with photorealistic style images.} We compare the stylized results of 5 scenes on Local Light Field Fusion(LLFF) ~\cite{mildenhall2019local} dataset. Our method stylizes scenes with more precise geometry and competitive stylization quality.}
	\label{fig:st_c_r}
	\vspace{-4mm}
\end{figure*}

\begin{figure*}[htbp]
	\centering
	\includegraphics[width=1.\linewidth]{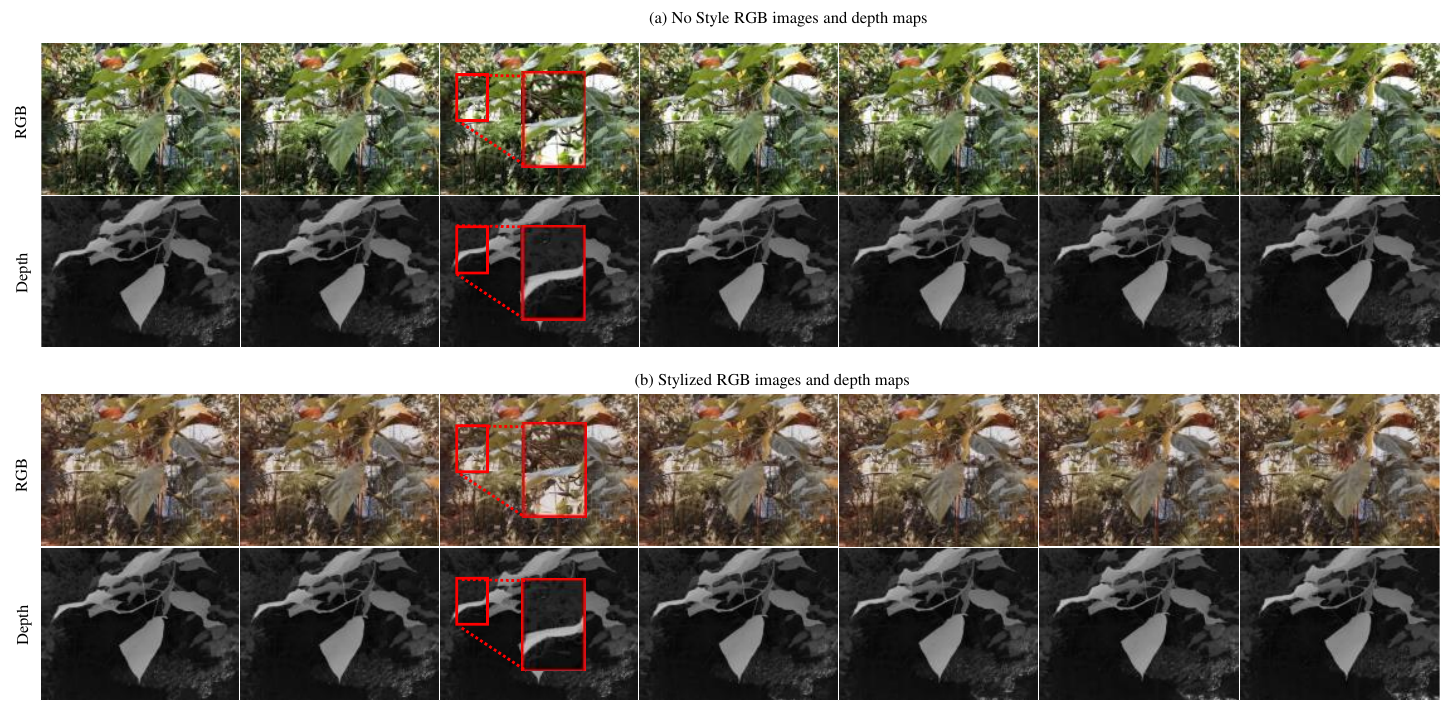}
	\caption{\textbf{Qualitative comparisons with no style multi-view images and stylized multi-view images.} The upper is the results without photorealistic style transfer, and the lower is the results with photorealistic style transfer from our method.}
	\label{fig:qvideo}
	\vspace{-4mm}
\end{figure*}

\section{Experiments}
\label{sec:experiment}

We have done the qualitative and quantitative evaluation tests for our method and also comparisons with the state-of-the-art stylization methods for video and 3D scenes, respectively. In our geometric training stage, we use the Adam optimizer with a batch size of 8,192 rays to optimize the scene representations for 20k iterations; in our style training stage, we use the Adam optimizer with a batch size of 10,000 rays to optimize the scene style representation for 200k iterations. The base learning rates are 0.1 for all voxel grids and $10^{-3}$ for RGBNet, HyperLinear, and HyperNet. We test our method on two types of datasets: NeRF-Synthetic datasets~\cite{mildenhall2020nerf} and  Local Light Field Fusion(LLFF) datasets~\cite{mildenhall2019local}. On the other hand, we use images in the MS-COCO~\cite{2014Microsoft} as the reference style images in the style training stage. All experiments are performed on a single NVIDIA TITAN RTX GPU. 

\subsection{Qualitative Results}
    
    \noindent\textbf{Photorealistic style transfers with artistic style images.} In Fig.~\ref{fig:st_c_a}, we qualitatively compare the photorealistic style transfer results generated by INS~\cite{fan2022unified}, Style3D~\cite{chiang2022stylizing}, Perceptual Loss~\cite{johnson2016perceptual}, MCCNet~\cite{deng:2020:arbitrary}, ReReVST~\cite{wang2020consistent} and ours. Although INS has targeted training for the style image in the stylized scene, its geometric representation is still far from our results. Other results in the absence of good geometry and the loss of precision, which further damages the stylization results. For example, the edge of the chair is not clear enough, and even one leg of the chair cannot be seen in the result of Style3D. At the same time, the artifacts of other methods are also severe. In contrast, our approach retains a clear geometric representation and can migrate a more realistic style from the style image, thereby changing the color in the scene.
  
     \noindent\textbf{Photorealistic style transfer with photorealistic style images.} In Fig.~\ref{fig:st_c_r}, we qualitatively compare the photorealistic style transfer results generated by ARF~\cite{zhang2022arf}, AdaIN~\cite{huang2017arbitrary},  MCCNet~\cite{deng:2020:arbitrary}, ReReVST~\cite{wang2020consistent} and ours with photorealistic style images. According to the default configuration of ARF, we retrained scenes with different realistic style images. MCCNet~\cite{deng:2020:arbitrary} and ReReVST~\cite{wang2020consistent} are two state-of-the-art video stylization methods. We should point out that ARF needs to retrain the scene according to the style image when rendering a new style scene, but our method does not require retraining. Instead, we can get a stylized scene by inputting the embedded features of the new style image into the network during rendering. We can see from the results that ARF will disorderly integrate the visual features in the style image into the scene when stylizing a new scene. However, our results perfectly transfer the color information according to the scene's style only, preserving the scene's geometric features to the greatest extent.

    \noindent\textbf{Video stylization.} In Fig.~\ref{fig:qvideo}, we compare our results in multiple views with and without photorealistic stylization. The results show our method almost has no effect on the depth value except the color of RGB. This is because we separate geometry training from style training and fix the voxel grid representing geometric features during style training stage.

\subsection{Quantitative Results}

\noindent\textbf{Consistency Measurement.} 

We measure the short and long-term consistency using the warped LPIPS metric~\cite{zhang2018unreasonable}. A view $v$ is warped with the depth expectation estimated by the learning from our geometric training. We use the measurement implemented from ~\cite{lai2018learning}. The consistency score formulates as:

\begin{equation}
    E(O_i, O_j) = LPIPS(O_i, M_{i,j}, W_{i,j}(O_j))
\end{equation}
where $W$ is the warping function and $M$ is the warping mask. When calculating the average distance across spatial dimensions in ~\cite{zhang2018unreasonable}, only pixels within the mask $M_{i,j}$ are taken. We compute the evaluation values on 5 scenes in the LLFF~\cite{mildenhall2019local} dataset, using 20 pairs of views for each scene. We use every two adjacent novel views ($O_{i},O_{i+1}$) and view pairs of gap 5 ($O_{i},O_{i+5}$) for short and long-range consistency calculation. The comparisons of short and long-range consistency are shown in Tab.~\ref{tab:short} and Tab.~\ref{tab:long}, respectively. Our method outperforms other methods by a significant margin.

\begin{table}
\centering
\caption{\textbf{Short-range consistency.} We compare the short-range consistency using warping error($\downarrow$). {\textbf{Best}} {results are highlighted.}}\label{tab:short}
\resizebox{\columnwidth}{!}{
\begin{tabular}{l|ccccc|c}

\hline													
Method	&	Fern	&	Flower	&	Horns	&	Orchids	&	Trex	&	Average	\\
\hline													
AdaIN	&	0.0051 	&	0.0033 	&	0.0055 	&	0.0066 	&	0.0041 	&	0.0049 	\\
MCCNet	&	0.0038 	&	0.0022 	&	0.0039 	&	0.0044 	&	0.0027 	&	0.0034 	\\
ReReVST	&	0.0011 	&	0.0007 	&	0.0011 	&	0.0019 	&	0.0009 	&	0.0011 	\\
ARF	&	0.0010 	&	0.0006 	&	0.0013 	&	0.0022 	&	0.0015 	&	0.0013 	\\
\hline													
Ours	&	\textbf{0.0005}	&	\textbf{0.0001}	&	\textbf{0.0003}	&	\textbf{0.0009}	&	\textbf{0.0003}	&	\textbf{0.0004}	\\
\hline
\end{tabular}
}
	\vspace{-2mm}
\end{table}

\begin{table}
\centering
\caption{\textbf{Long-range consistency.} We compare the long-range consistency using warping error($\downarrow$).  {\textbf{Best}} {results are highlighted.}}\label{tab:long}
\resizebox{\columnwidth}{!}{
\begin{tabular}{l|ccccc|c}
	\hline
Method	&	Fern	&	Flower	&	Horns	&	Orchids	&	Trex	&	Average	\\
\hline													
AdaIN	&	0.0087 	&	0.0063 	&	0.0097 	&	0.0100 	&	0.0078 	&	0.0085 	\\
MCCNet	&	0.0070 	&	0.0042 	&	0.0078 	&	0.0074 	&	0.0058 	&	0.0065 	\\
ReReVST	&	0.0035 	&	0.0025 	&	0.0035 	&	0.0053 	&	0.0024 	&	0.0035 	\\
ARF	&	0.0042 	&	0.0027 	&	0.0053 	&	0.0075 	&	0.0051 	&	0.0050 	\\
\hline													
Ours	&	\textbf{0.0024}	&	\textbf{0.0009}	&	\textbf{0.0020}	&	\textbf{0.0032}	&	\textbf{0.0015}	&	\textbf{0.0020}	\\
\hline

\end{tabular}}
	\vspace{-5mm}
\end{table}

\begin{figure}[htbp]
	\vspace{1mm}
	\centering
	\includegraphics[width=0.75\linewidth]{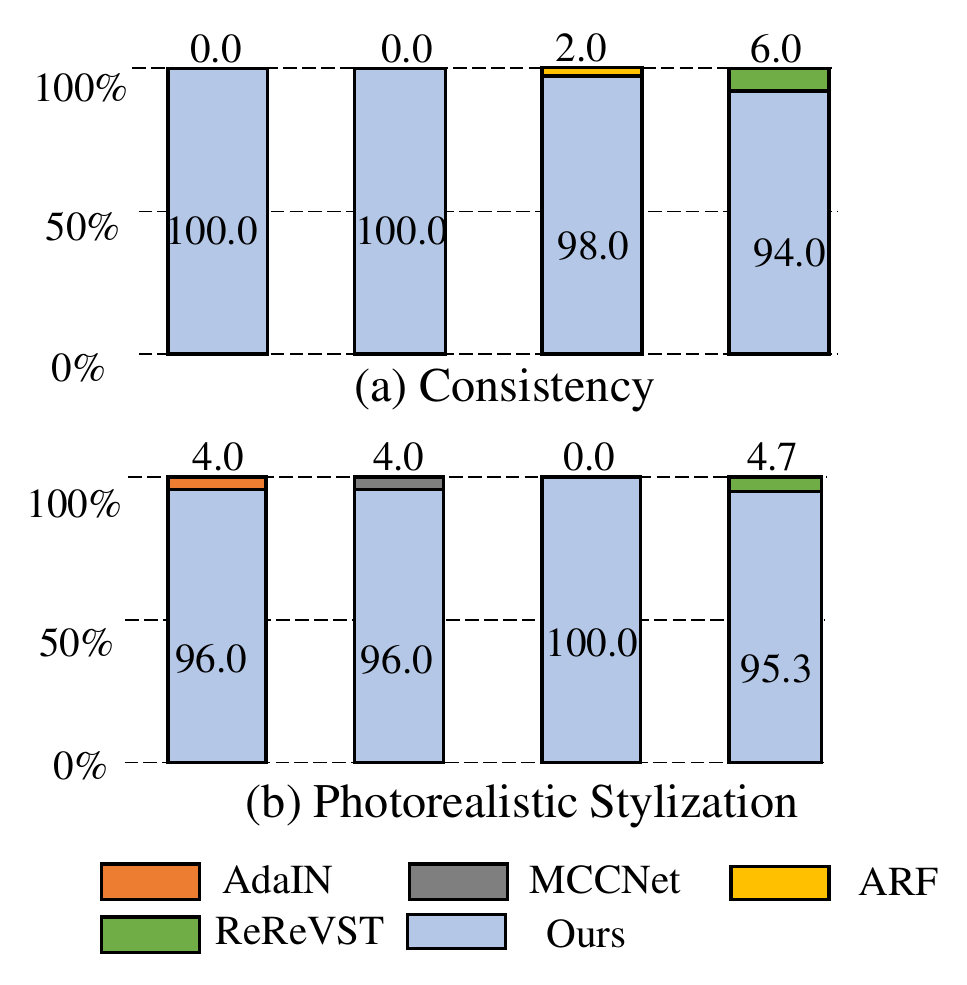}
	\caption{\textbf{User study.} We record the user preference in the form of boxplot. Our results win more preferences both in the photorealistic stylization and consistency quality.}
\label{fig:user_study}
	\vspace{-4mm}
\end{figure}

\noindent\textbf{User study.} A user study is conducted to compare our method's stylization and consistent quality with other state-of-the-art methods. We stylize ten series of views of the 3D scenes in the LLFF~\cite{mildenhall2019local} dataset, using different methods~\cite{deng:2020:arbitrary},~\cite{wang2020consistent},~\cite{huang2021learning} and invite 30 participants (including 25 males, 5 females, aged from 20 to 43). First, we showed the participants a style image, two stylized videos generated by our method, and a random compared method. Then we counted the participants their votes for the video in two evaluating indicators, quality of the stylized results and whether to keep the consistency. We collected 600 votes for each evaluating indicator and presented the result in Fig.~\ref{fig:user_study} in the form of the boxplot. Our scores stand out from other methods in photorealistic stylization quality and consistency.
\begin{figure}[htbp]
	\centering
	\includegraphics[width=0.95\linewidth]{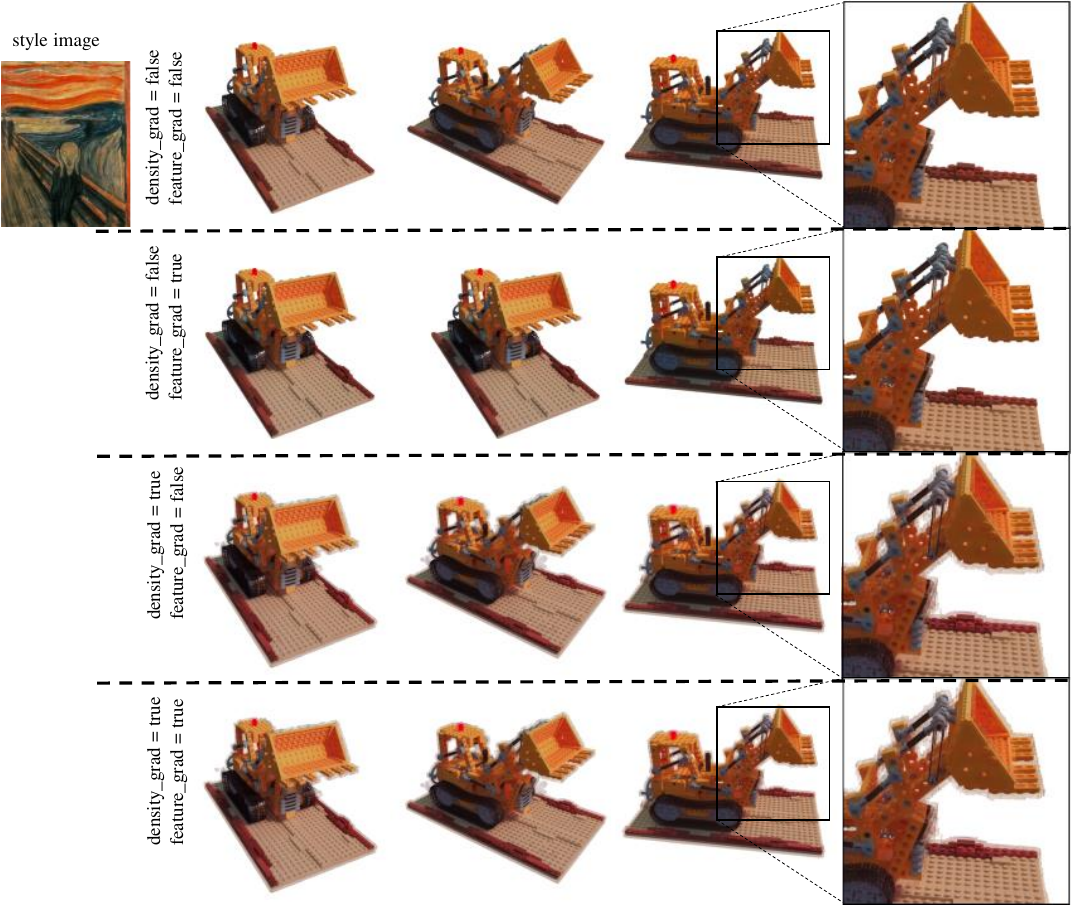}
	\caption{\textbf{The impact of voxel grid gradient propagation in style training stage.} $L_d$ clusters latent codes of the same style and avoids the artifacts in test results.}
	\label{fig:ab_grad}
	\vspace{-5mm}
\end{figure}

\subsection{Ablation Study}

\noindent\textbf{The impact of voxel grid gradient propagation in the style training stage.} We believe that the most critical performance of photorealistic style transfer is that the scene's color needs to be consistent with the reference style image, and the methods cannot change the geometric information of the scene. That is, photorealistic style transfer should not change the geometric shape of the scene. Based on this principle, we first trained the scene's geometry and then froze the parameters of the voxel grid for style training.
We explored the impact of voxel gradient propagation in the style training stage.
We try to freeze the parameters of the density voxel grid and feature voxel grid, respectively, in the process of gradient promotion and then compare the results of style transfer of the trained network. The result shown in Fig. \ref{fig:ab_grad}. $\operatorname{density\_grad=true}$ and $\operatorname{feature\_grad=true}$ indicate the parameters in density voxel grid and feature voxel grid not be frozen in style training, respectively. From the results, we can see that as long as we freeze the parameters of the density voxel grid in the style training stage, we can get a better photorealistic style transfer effect while keeping the geometric information of the scene.

\begin{figure}[htbp]
	\vspace{-1mm}
	\centering
	\includegraphics[width=0.95\linewidth]{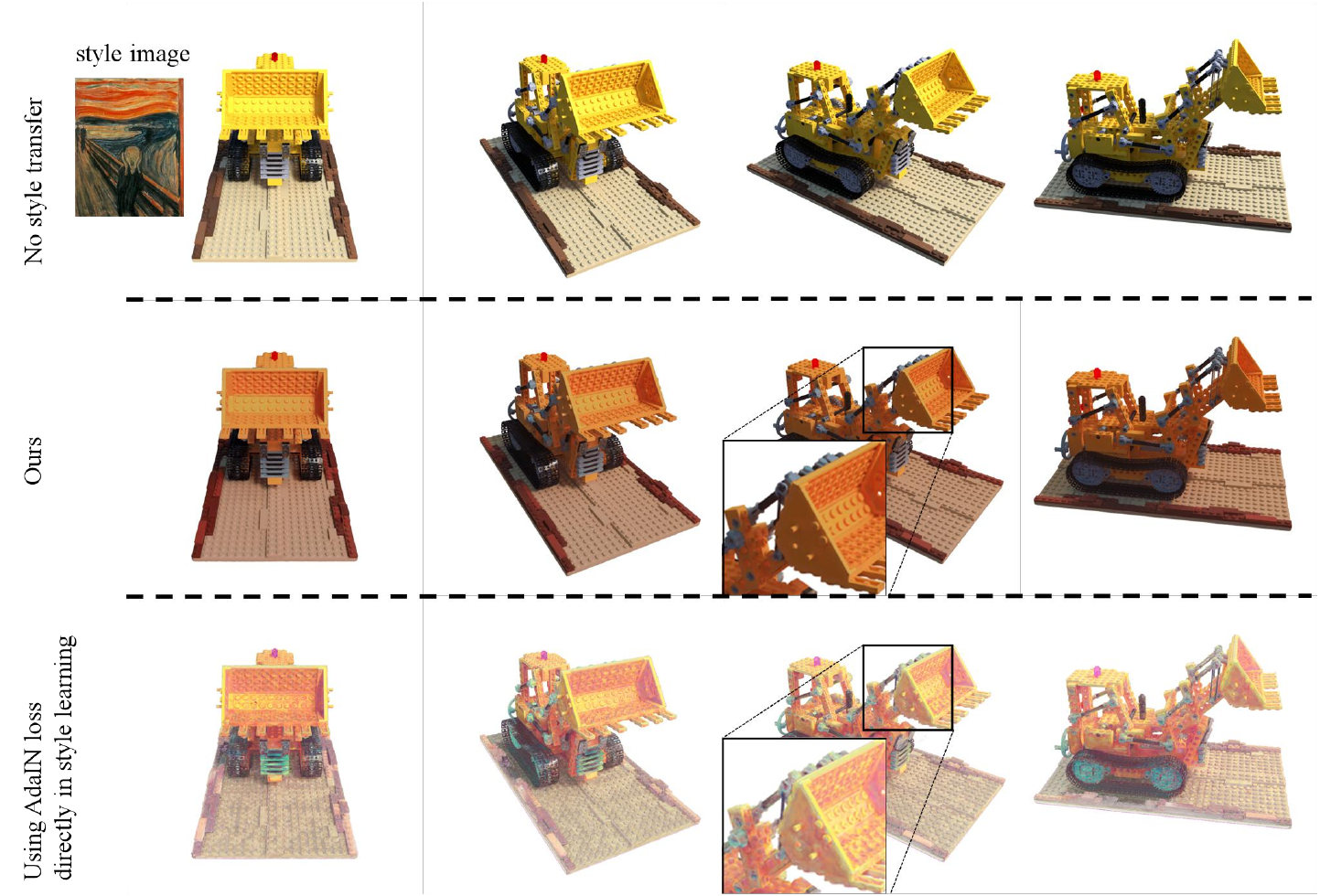}
	\caption{\textbf{The impact of using 2D photorealistic style network to constrain scene style.}}
	\label{fig:ab_loss}
	\vspace{-4mm}
\end{figure}

\noindent\textbf{The impact of using a 2D photorealistic style network to constrain scene style.} In our method, we design a virtual 2D photorealistic style transfer network, YUVStyleNet, which is used to generate photorealistic style images in the style training stage to constrain the style of the scene. This will significantly ensure the quality of photorealistic style transfer scene. To verify this, we directly use AdaIN as a loss to constrain the style training process. As a result, the direct use of AdaIN constraints is more blurred in the generated novel view images than in our method, as shown in Fig. \ref{fig:ab_loss}.

\begin{figure}[htbp]
	\centering
	\includegraphics[width=1.\linewidth]{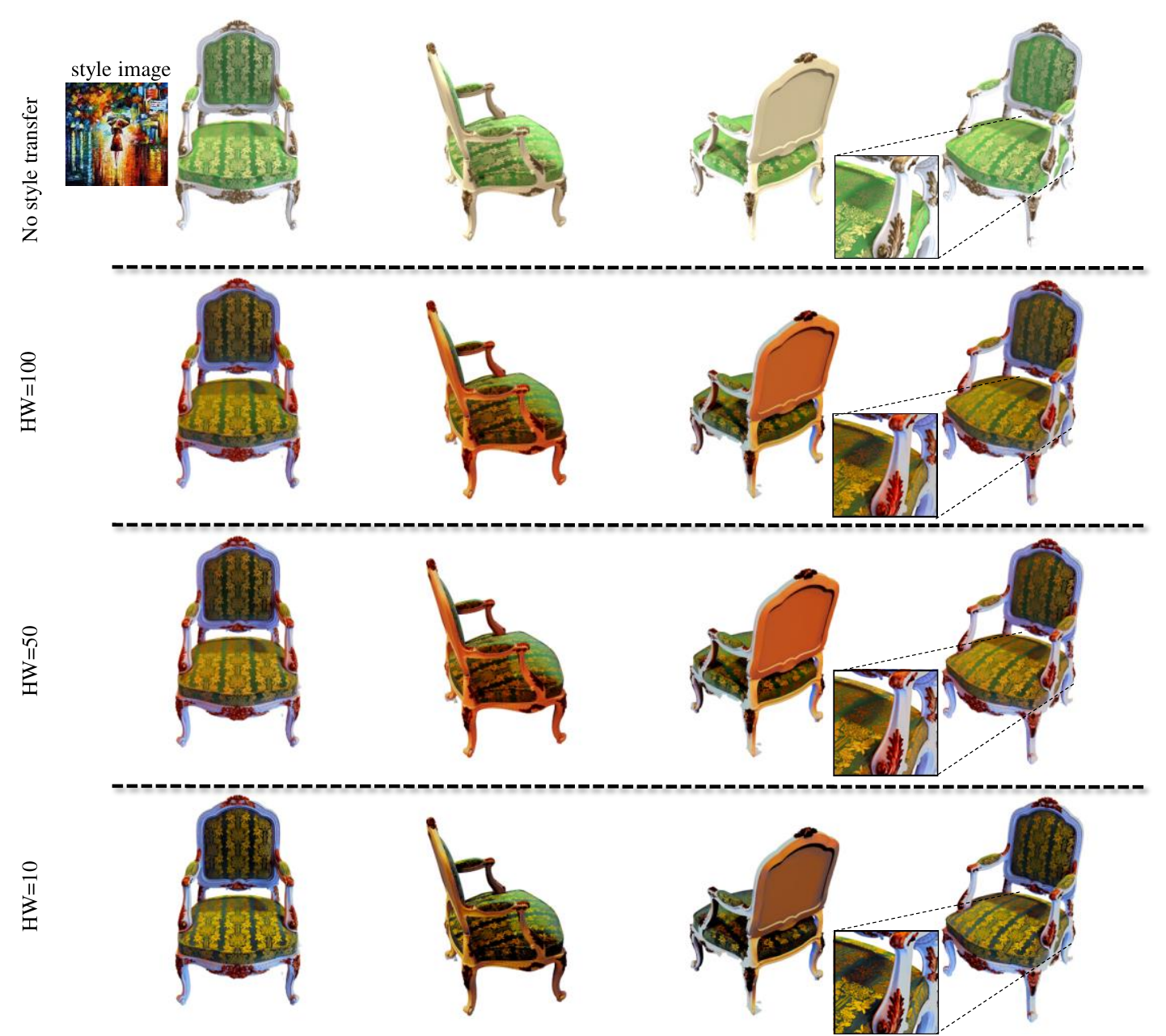}
	\caption{\textbf{Ablation study on a batch size of rays in style training.} We compare the results with different batches of rays in style training. HW=10 indicate the batch size is $10 \times 10 = 100$ of rays.}
	\label{fig:ab_size}
	\vspace{-4mm}
\end{figure}

\begin{figure}[htbp]
	\centering
	\includegraphics[width=1.\linewidth]{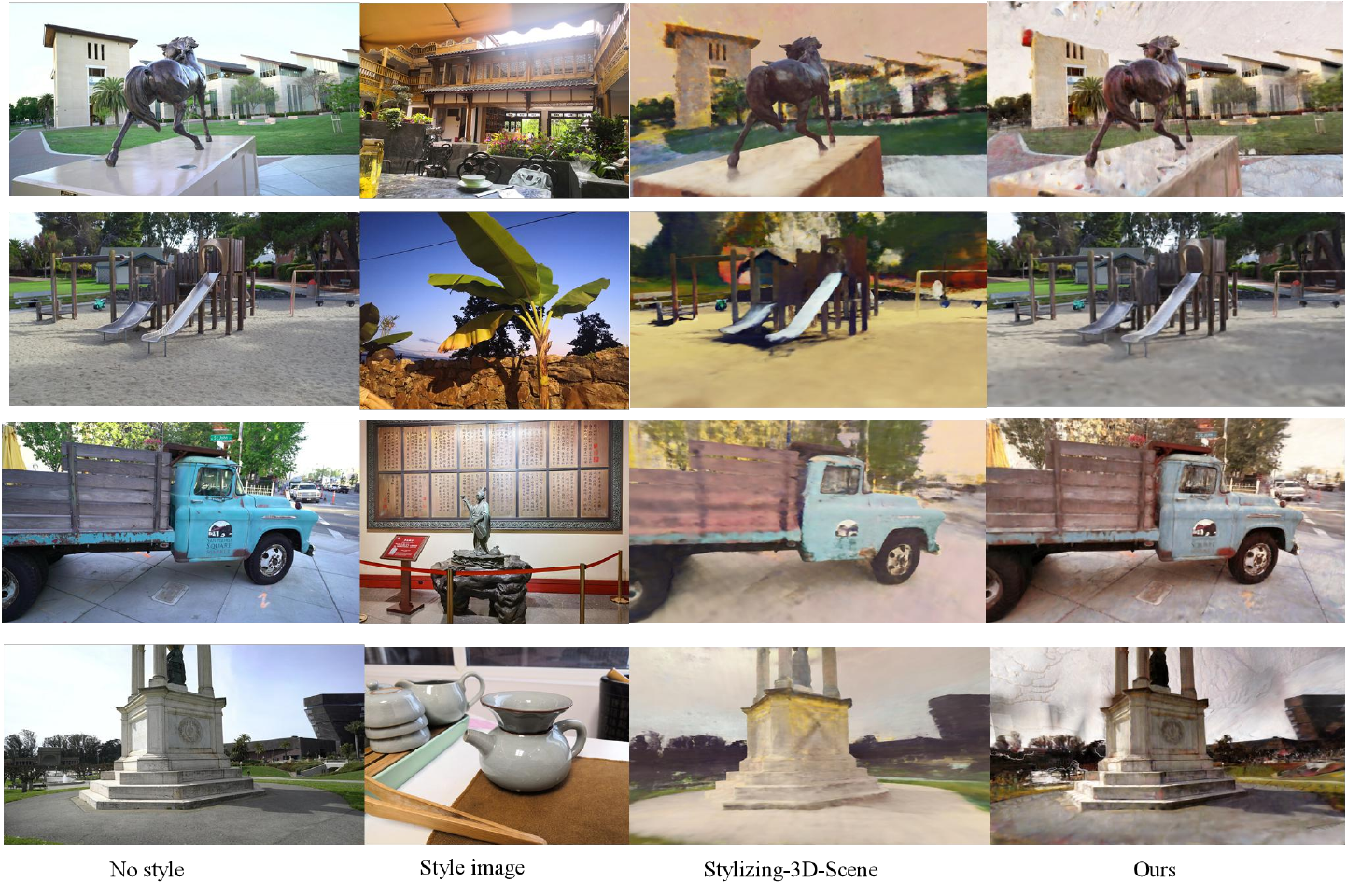}
	\caption{\textbf{Comparisions on Tanks and Temples ~\cite{knapitsch2017tanks} datasets.} We compare the results on the large scale of sense datasets, artifacts in the results may exist.}
	\label{fig:ab_large}
	\vspace{-4mm}
\end{figure}

\noindent\textbf{The impact of a batch size of rays in style training.} We conduct stylization constraints in the style training stage by processing 2D style images and small batches of novel view images. When the batch size is larger, the novel view of the captured scene will be larger, with higher global constraints. On the contrary, it will be closer to local constraints. We studied this in Fig.~\ref{fig:ab_size}. HW is the space through which a rectangular batch of rays passes. For example, if HW = 10, the batch size of the rays is $10 \times 10=100$. We can see from the results that the change from 10 to 100 causes the change of color, but the overall impact is negligible.

\noindent\textbf{Limitations.} The quality of the photorealistic stylization results is limited by the geometric training stage. We use a voxel grid to represent the geometric of the scene. When the scene to be represented is large enough, this method will consume huge storage space. Therefore, the maximum value of the size of the voxel grid is limited, so the method cannot reconstruct some large scenes well enough. This also affects the final photorealistic transfer results. Fig. ~\ref{fig:ab_large} shows the results of a large scene dataset. It can be seen that our method has artifacts in the sky.
\section{Conclusion}
We present a universal photorealistic style transfer method with neural radiance fields for the 3D scene. We directly reconstruct the geometric representation through the voxel grid and then introduce the features of different 2D style images for scene style control in the style training stage. To achieve this, we use a hyper network to control the weights. Further, we use the pre-trained 2D photorealistic style network to perform photorealistic style transfer on the input style image and the novel view image of the scene to constrain the training of scene photorealistic style transfer. Our method outperforms state-of-the-art methods both in terms of visual quality and consistency. However, our direct optimization of scene geometry via voxel grid has limitations in large 3D scenes. In the future, we will explore the problem of photorealistic style transfer in large scenes. At the same time, we will focus on exploring the use of neural radiation fields to solve the problem of color consistency in different scenes.

\section*{Acknowledgement}

\noindent This work was supported in part by the National Natural Science Foundation of China under Grant No. 62172061; National Key R\&D Program of China under Grant No. 2020YFB1711800 and 2020YFB1707900. We are grateful to thank the support from Peng Cheng Laboratory. We sincerely appreciate all participants in the user study.

{\small
	\bibliographystyle{ieee_fullname}
	\bibliography{camera_ready}
}

\appendix
\newpage

\section{Supplementary Material}
\label{sec.details}
\subsection{Detailed Configuration of Neural Network }

Tab. ~\ref{tab:RGBNet}, ~\ref{tab:HyperNet} and ~\ref{tab:HyperLiner} are the detailed configurations of the neural networks used in our framework of universal photorealistic style transfer which shown in Fig.\ref{fig:framework}. In these tables, $\operatorname{OP}$ refers to  $\operatorname{Operation}$, $\operatorname{IN}$ refers to  number of the $\operatorname{Input}$ channels of the features,  $\operatorname{OUT}$ refers to  number of the $\operatorname{Output}$ channels of the features and $\operatorname{ACT}$ refers to the $\operatorname{Activation}$ function.  Our HyperNet takes the features of the style image as input to control the weight of HyperLiner, so as to change the color style of the scene. Therefore, the number of output channels in HyperNet is determined according to the number of output and input channels in HyperLiner. For example, the number of output channel of the $\operatorname{Hyper0}$ layer in HyperNet is $5120$, which is determined by the number of input channel $\operatorname{IN}$ ($39$) and the number of output channel $\operatorname{OUT}$ ($128$) of the $0$ layer in hyperliner.

\begin{table}[h]\tiny
	\centering
	\caption{{Detailed configuration of RGBNet.} }\label{tab:RGBNet}
	\resizebox{\columnwidth}{!}{
		\begin{tabular}{l|cccccc}
			
			\hline									
			Layers	&	$\operatorname{OP}$	&&&	$\operatorname{IN}$	&	$\operatorname{OUT}$	&	$\operatorname{ACT}$	\\
			\hline							
			0	&	Linear	&&&	39	&	128	&	ReLU	\\
			
			1	&	Linear	&&&	128	&	128	&	ReLU	\\
			
			2	&	Linear	&&&	128	&	3	&		\\
			\hline

		\end{tabular}
	}
	\vspace{-2mm}
\end{table}

\begin{table}[h]\tiny
	\centering
	\caption{{Detailed configuration of HyperNet.} }\label{tab:HyperNet}
	\resizebox{\columnwidth}{!}{
		\begin{tabular}{l|cccccc}
			
\hline									
Layers	&	$\operatorname{OP}$	&	$\operatorname{IN}$	&	$\operatorname{OUT}$	&	$\operatorname{ACT}$	\\
\hline									
\multirow{3}{*}{Hyper0}	&	Linear	&	512	&	64	&	ReLU	\\
&	Linear	&	64	&	64	&	ReLU	\\
&	Linear	&	64	&	5120(39*128+128)	&	ReLU	\\
\hline									
\multirow{3}{*}{Hyper1}	&	Linear	&	512	&	64	&	ReLU	\\
&	Linear	&	64	&	64	&	ReLU	\\
&	Linear	&	64	&	16512(128*128+128)	&	ReLU	\\
\hline									
\multirow{3}{*}{Hyper2}	&	Linear	&	512	&	64	&	ReLU	\\
&	Linear	&	64	&	64	&	ReLU	\\
&	Linear	&	64	&	16512(128*128+128)	&	ReLU	\\
\hline									
\multirow{3}{*}{Hyper3}	&	Linear	&	512	&	64	&	ReLU	\\
&	Linear	&	64	&	64	&	ReLU	\\
&	Linear	&	64	&	8256(64*128+64)	&	ReLU	\\
\hline									
\multirow{3}{*}{Hyper4}	&	Linear	&	512	&	64	&	ReLU	\\
&	Linear	&	64	&	64	&	ReLU	\\
&	Linear	&	64	&	2535(64*39+64)	&	ReLU	\\
\hline									

		\end{tabular}
	}
	\vspace{-2mm}
\end{table}

\begin{table}[h]\tiny
	\centering
	\caption{{Detailed configuration of HyperLiner.} }\label{tab:HyperLiner}
	\resizebox{\columnwidth}{!}{
		\begin{tabular}{l|cccccc}
			
\hline									
Layers	&	$\operatorname{OP}$	&&	$\operatorname{IN}$	&	$\operatorname{OUT}$	&	$\operatorname{ACT}$	\\
\hline									
0	&	BatchLinear	&&	39	&	128	&	ReLU	\\
1	&	BatchLinear	&&	128	&	128	&	ReLU	\\
2	&	BatchLinear	&&	128	&	128	&	ReLU	\\
3	&	BatchLinear	&&	128	&	64	&	ReLU	\\
4	&	BatchLinear	&&	64	&	39	&	ReLU	\\
\hline

		\end{tabular}
	}
	\vspace{-2mm}
\end{table}

Tab. ~\ref{tab:ConvolutionalYUVStyleNet} and ~\ref{tab:sbYUVStyleNet} show the detailed configurations of our 2D photorealistic stylization framework YUVStyleNet which shown in Fig.~\ref{fig:YUVStyleNet}. $\operatorname{k}$ refers to the $\operatorname{kernel}$ size of the convolution and $\operatorname{s}$ refers to the $\operatorname{stride}$ size.  

\begin{table}[h]\tiny
	\centering
	\caption{{Detailed configuration of Convolutional Network in YUVStyleNet.} }\label{tab:ConvolutionalYUVStyleNet}
	\resizebox{\columnwidth}{!}{
		\begin{tabular}{l|cccccccccc}
			\hline													
			Layers	&	$\operatorname{OP}$	&	$\operatorname{IN}$	&	$\operatorname{OUT}$	&	$\operatorname{k}$	&	$\operatorname{s}$	&	$\operatorname{ACT}$	\\
			\hline													
			\multirow{2}{*}{Conv5}	&	Conv2d	&	512	&	16	&	3	&	1	&	LeakyReLU	\\
			&	Conv2d	&	16	&	256	&	3	&	1	&	Sigmoid	\\
			\hline	
			\multirow{2}{*}{Conv4}	&	Conv2d	&	512	&	16	&	3	&	1	&	LeakyReLU	\\
			&	Conv2d	&	16	&	128	&	3	&	1	&	Sigmoid	\\
			\hline	
			\multirow{2}{*}{Conv3}	&	Conv2d	&	256	&	16	&	3	&	1	&	LeakyReLU	\\
			&	Conv2d	&	16	&	64	&	3	&	1	&	Sigmoid	\\
			\hline	
			\multirow{2}{*}{Conv2}	&	Conv2d	&	128	&	16	&	3	&	1	&	LeakyReLU	\\
			&	Conv2d	&	16	&	3	&	3	&	1	&	Sigmoid	\\
			\hline	
			\multirow{2}{*}{Conv1}	&	Conv2d	&	6	&	16	&	3	&	1	&	LeakyReLU	\\
			&	Conv2d	&	16	&	3	&	3	&	1	&	Sigmoid	\\
			\hline

		\end{tabular}
	}
	\vspace{-2mm}
\end{table}

\begin{table}[h]\tiny
	\centering
	\caption{Detailed configuration of Splatting Blocks in YUVStyleNet.}\label{tab:sbYUVStyleNet}
	\resizebox{\columnwidth}{!}{
		\begin{tabular}{l|cccccccc}
\hline											
Layers	&	$\operatorname{OP}$	&	$\operatorname{IN}$	&	$\operatorname{OUT}$	&	$\operatorname{k}$	&	$\operatorname{s}$	\\
\hline											
\multirow{7}{*}{SB1}	&	\multicolumn{5}{c}{low-res style}					\\
						&	ReflectionPad2d						&	3	&	3	&	/	&	/	\\
						&	Conv2d								&	3	&	3	&	3	&	1	\\
						&	\multicolumn{5}{c}{low-res content}			\\
						&	ReflectionPad2d						&	3	&	3	&	/	&	/	\\
						&	Conv2d								&	3	&	3	&	3	&	1	\\
						&	\multicolumn{5}{c}{AdaIN}			\\
\hline	
\multirow{7}{*}{SB2}	&	\multicolumn{5}{c}{SF4}\\
						&	ReflectionPad2d						&	64	&	64	&	/	&	/	\\
						&	Conv2d								&	64	&	64	&	3	&	1	\\
						&	\multicolumn{5}{c}{CF4}\\
						&	ReflectionPad2d						&	64	&	64	&	/	&	/	\\
						&	Conv2d								&	64	&	64	&	3	&	1	\\
						&\multicolumn{5}{c}{AdaIN}			\\
\hline	
\multirow{7}{*}{SB3}	&	\multicolumn{5}{c}{SF3}\\
						&	ReflectionPad2d						&	128	&	128	&	/	&	/	\\
						&	Conv2d								&	128	&	128	&	3	&	1	\\
						&	\multicolumn{5}{c}{CF3}\\
						&	ReflectionPad2d						&	128	&	128	&	/	&	/	\\
						&	Conv2d								&	128	&	128	&	3	&	1	\\
						&\multicolumn{5}{c}{AdaIN}			\\
\hline	
\multirow{7}{*}{SB4}	&	\multicolumn{5}{c}{SF2}\\
						&	ReflectionPad2d						&	256	&	256	&	/	&	/	\\
						&	Conv2d								&	256	&	256	&	3	&	1	\\
						&	\multicolumn{5}{c}{CF2}\\
						&	ReflectionPad2d						&	256	&	256	&	/	&	/	\\
						&	Conv2d								&	256	&	256	&	3	&	1	\\
						&	\multicolumn{5}{c}{AdaIN}			\\
\hline	
\multirow{7}{*}{SB5}	&	\multicolumn{5}{c}{SF1}\\
						&	ReflectionPad2d						&	512	&	512	&	/	&	/	\\
						&	Conv2d								&	512	&	512	&	3	&	1	\\
						&   \multicolumn{5}{c}{CF1}\\
						&	ReflectionPad2d						&	512	&	512	&	/	&	/	\\
						&	Conv2d								&	512	&	512	&	3	&	1	\\
						&	\multicolumn{5}{c}{AdaIN}			\\
\hline

		\end{tabular}
	}
	\vspace{-2mm}
\end{table}

In Tab.~\ref{tab:sbYUVStyleNet}, we use adaptive instance normalization (AdaIN)~\cite{huang2017arbitrary} to fuse $s$ feature and $c$ feature from the splatting block module. Specifically, let $s,c  \in \mathbb{R}^{C \times H \times W }$, then AdaIN is defined as:
\begin{equation} \label{eq:AdaIN}
	\operatorname{AdaIN}(c,s) = \sigma(s)\frac{c-\mu(c)}{\sigma(c)}+\mu(s)
\end{equation}
where $\mu(c)$ and $\sigma(c)$ (resp. $\mu(s)$ and $\sigma(s)$) are the mean and standard deviation of $c$ (resp.$s$) over its spatial dimension.

\begin{figure*}[ht]
	\centering
	\begin{tabular}{c@{\hspace{0.002\linewidth}}c@{\hspace{0.002\linewidth}}c@{\hspace{0.002\linewidth}}c@{\hspace{0.002\linewidth}}c@{\hspace{0.002\linewidth}}c}
		
		\includegraphics[width = .13\linewidth,height=.14\linewidth]{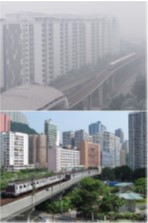} &
		\includegraphics[width = .22\linewidth,height=.14\linewidth]{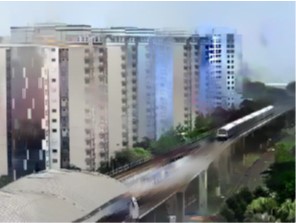} &
		\includegraphics[width = .22\linewidth,height=.14\linewidth]{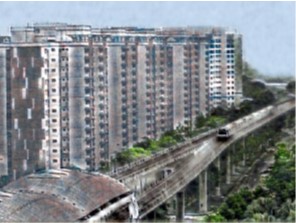} &
		\includegraphics[width = .22\linewidth,height=.14\linewidth]{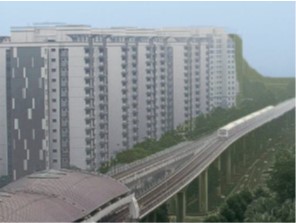} &
		\includegraphics[width = .22\linewidth,height=.14\linewidth]{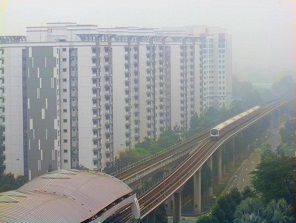} &\\ 
		\vspace{-0.1em}
		
		\includegraphics[width = .13\linewidth,height=.14\linewidth]{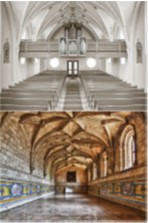} &
		\includegraphics[width = .22\linewidth,height=.14\linewidth]{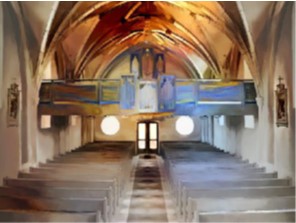} &
		\includegraphics[width = .22\linewidth,height=.14\linewidth]{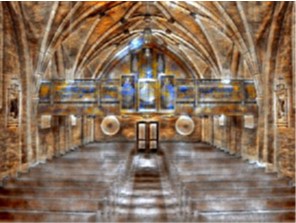} &
		\includegraphics[width = .22\linewidth,height=.14\linewidth]{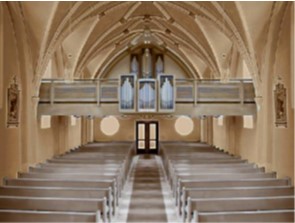} &
		\includegraphics[width = .22\linewidth,height=.14\linewidth]{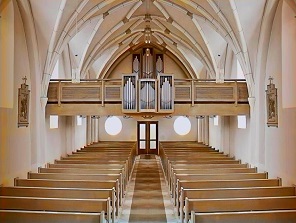} &\\
		\vspace{-0.1em}
		
		\includegraphics[width = .13\linewidth,height=.14\linewidth]{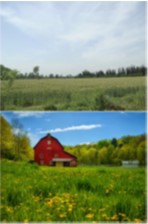} &
		\includegraphics[width = .22\linewidth,height=.14\linewidth]{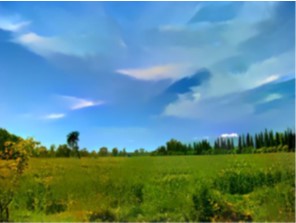} &
		\includegraphics[width = .22\linewidth,height=.14\linewidth]{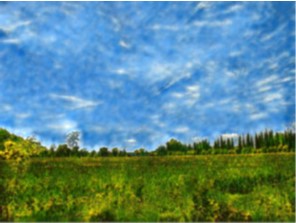} &
		\includegraphics[width = .22\linewidth,height=.14\linewidth]{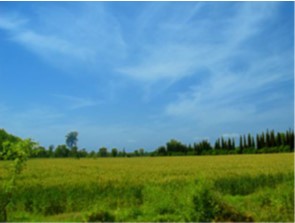} &
		\includegraphics[width = .22\linewidth,height=.14\linewidth]{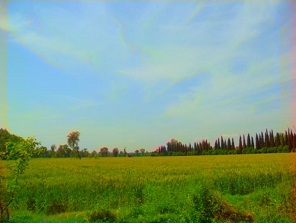} &\\
		\vspace{-0.1em}

		\includegraphics[width = .13\linewidth,height=.14\linewidth]{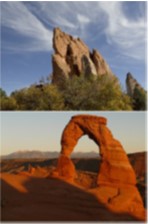} &
		\includegraphics[width = .22\linewidth,height=.14\linewidth]{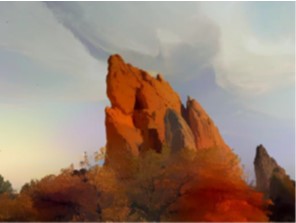} &
		\includegraphics[width = .22\linewidth,height=.14\linewidth]{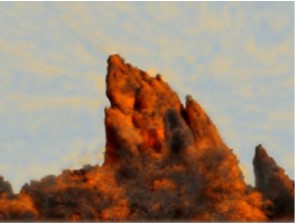} &
		\includegraphics[width = .22\linewidth,height=.14\linewidth]{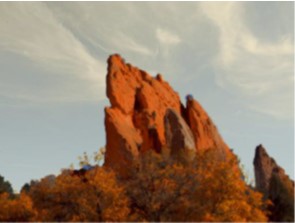} &
		\includegraphics[width = .22\linewidth,height=.14\linewidth]{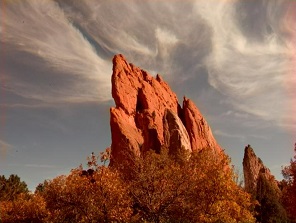} &\\
		\vspace{-0.1em}
		
		\includegraphics[width = .13\linewidth,height=.14\linewidth]{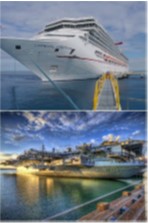} &
		\includegraphics[width = .22\linewidth,height=.14\linewidth]{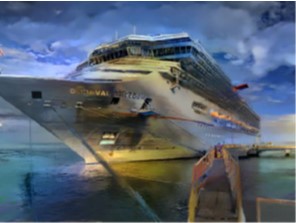} &
		\includegraphics[width = .22\linewidth,height=.14\linewidth]{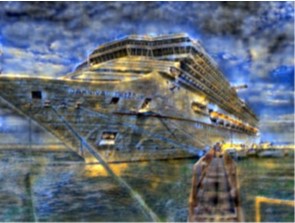} &
		\includegraphics[width = .22\linewidth,height=.14\linewidth]{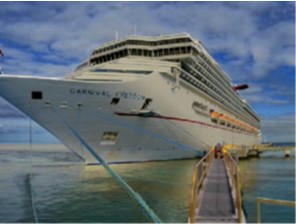} &
		\includegraphics[width = .22\linewidth,height=.14\linewidth]{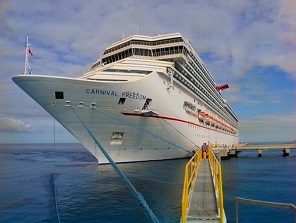} &\\
		\vspace{-0.1em}

		\includegraphics[width = .13\linewidth,height=.14\linewidth]{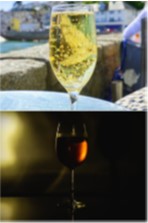} &
		\includegraphics[width = .22\linewidth,height=.14\linewidth]{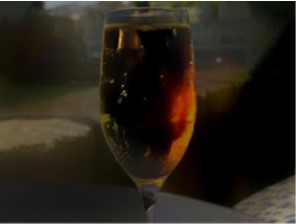} &
		\includegraphics[width = .22\linewidth,height=.14\linewidth]{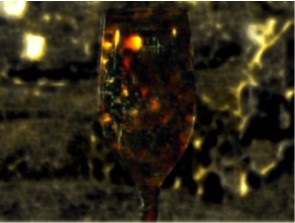} &
		\includegraphics[width = .22\linewidth,height=.14\linewidth]{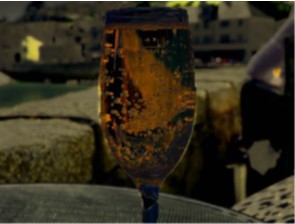} &
		\includegraphics[width = .22\linewidth,height=.14\linewidth]{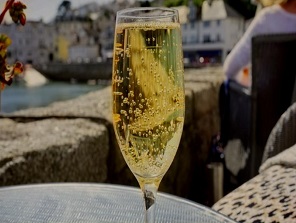} &\\
		\vspace{-0.1em}

		\vspace{-1em}
		Inputs & {\small PhotoWCT~\cite{li2018closed}} & LST~\cite{li2019learning} & WCT\textsuperscript{2}~\cite{yoo2019photorealistic} & YUVStyleNet(Ours) \\
	\end{tabular}
	\vspace{0.5em}
	\caption{{YUVStyleNet qualitative comparison for 2D photorealistic stylization}. Our method against three state of the art baselines on some challenging examples.}
	\label{fig:app_2d_qualitative}
\end{figure*}
\subsection{Additional Visual Results }
We show more results, including comparing 2D photorealistic stylization methods and more stylized results of 3D scenes. Fig. ~\ref{fig:app_2d_qualitative} is a comparison between our designed 2D photorealistic stylization method and other 2D photorealistic stylization methods. Our results have better visual quality than others.

\begin{figure*}[htbp]
	\centering
	\includegraphics[width=1.\linewidth]{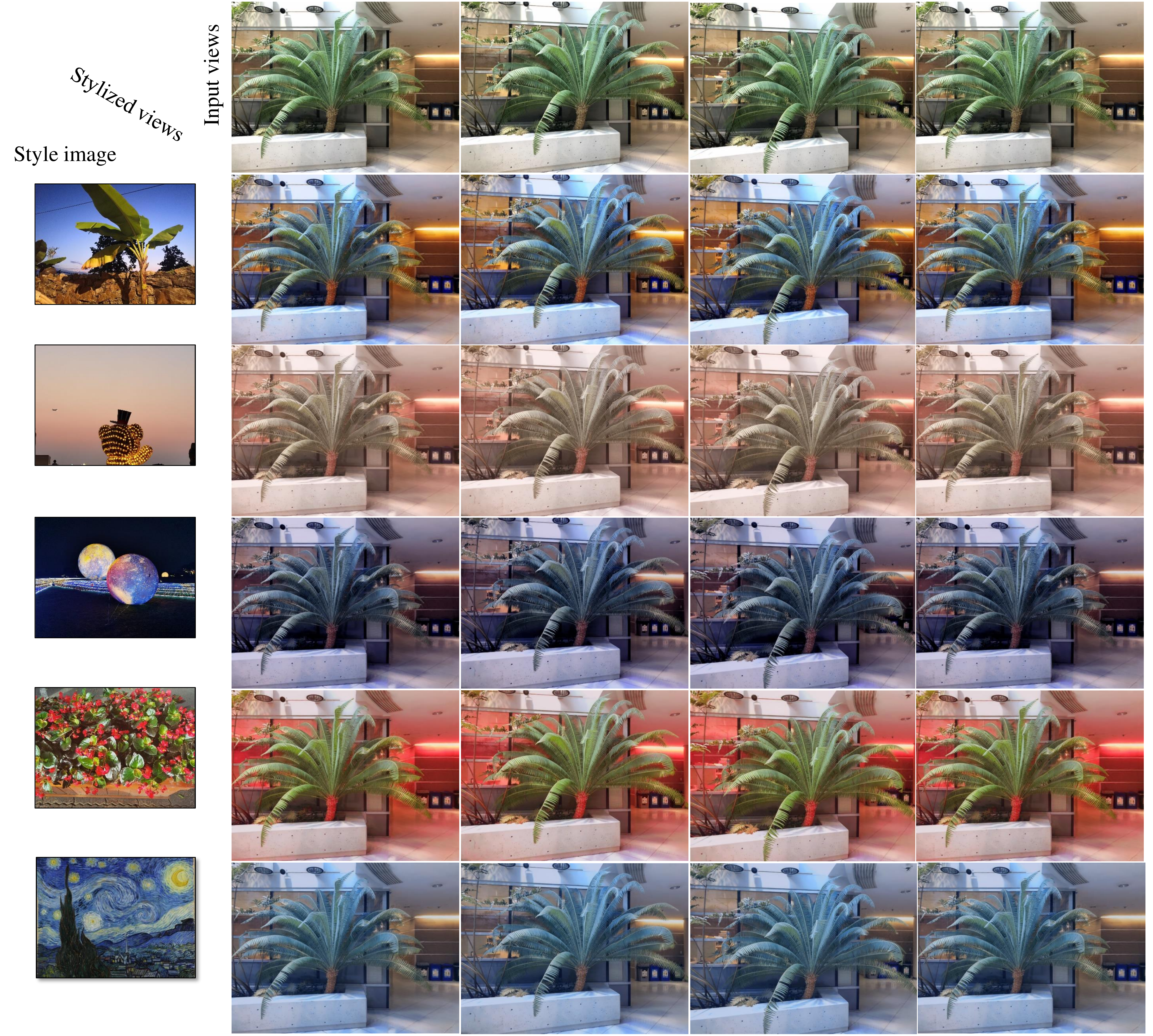}
	\caption{Photorealistic stylization results with the $\operatorname{fern}$ 3D scene on LLFF dataset.}
	\label{fig:app_s_fern}
	\vspace{-4mm}
\end{figure*}

\begin{figure*}[htbp]
	\centering
	\includegraphics[width=1.\linewidth]{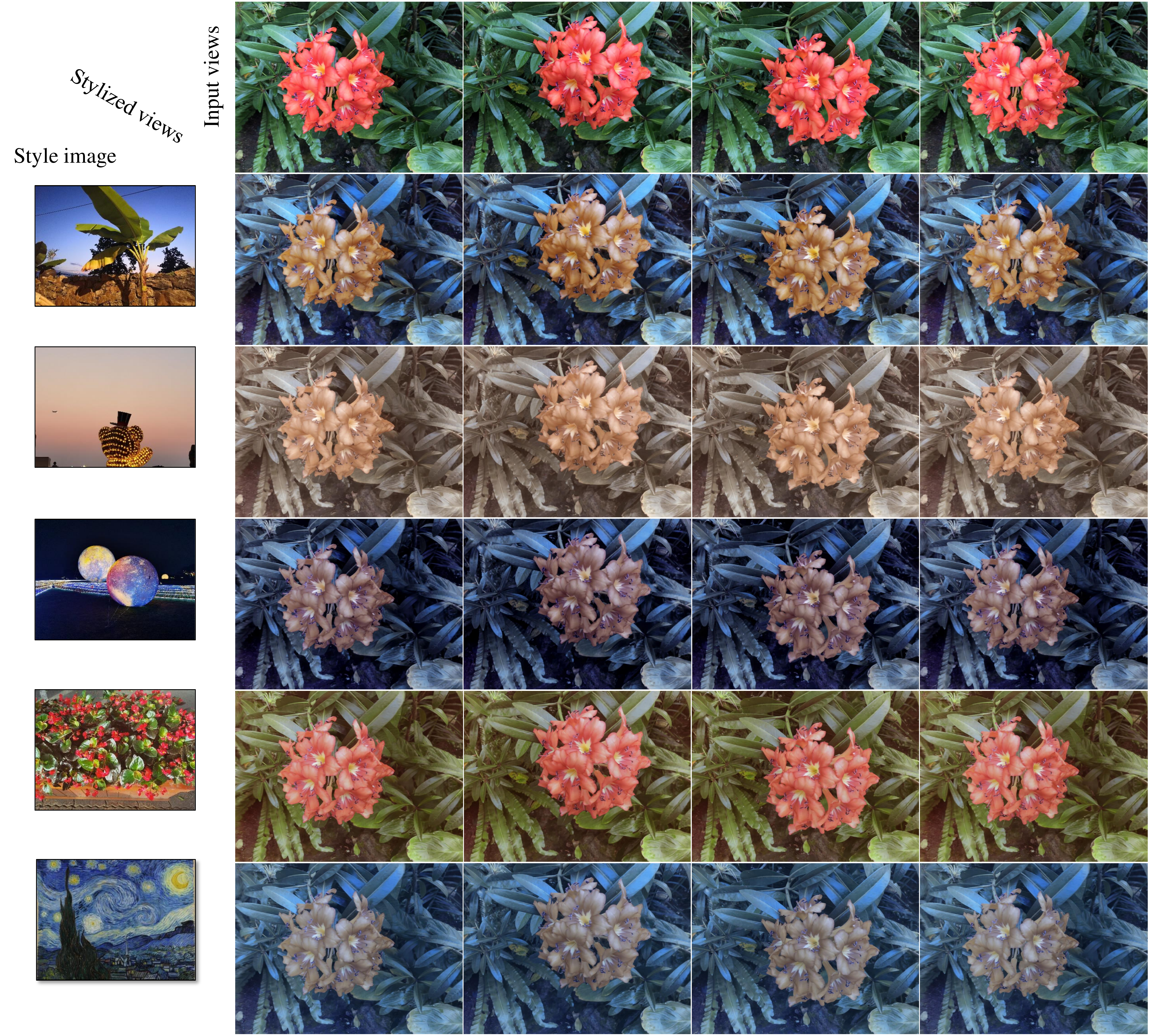}
	\caption{Photorealistic stylization results with the $\operatorname{flower}$ 3D scene on LLFF dataset.}
	\label{fig:app_s_flower}
	\vspace{-4mm}
\end{figure*}

\begin{figure*}[htbp]
	\centering
	\includegraphics[width=1.\linewidth]{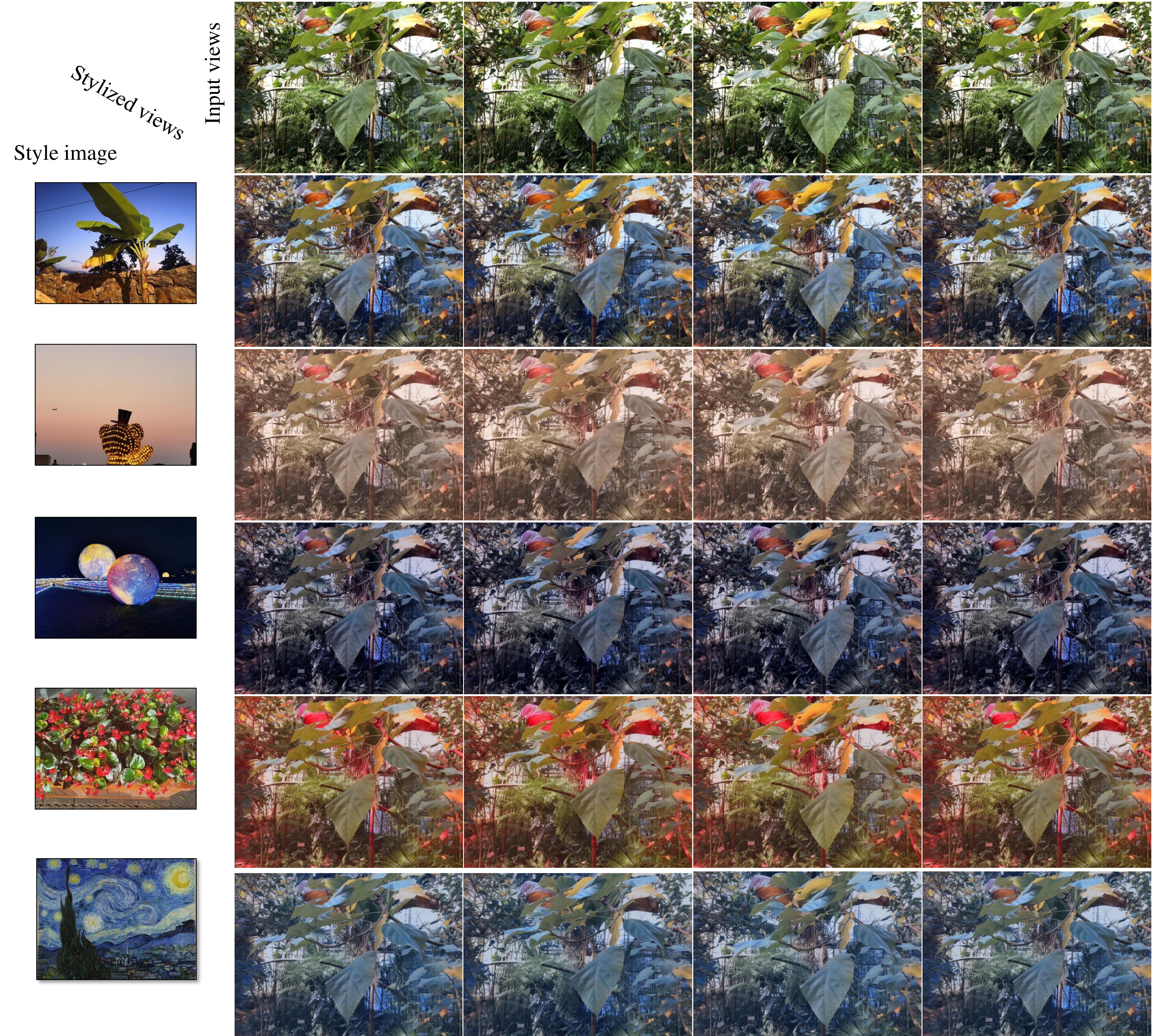}
	\caption{Photorealistic stylization results with the $\operatorname{leaves}$ 3D scene on LLFF dataset.}
	\label{fig:app_s_leaves}
	\vspace{-4mm}
\end{figure*}

\begin{figure*}[htbp]
	\centering
	\includegraphics[width=1.\linewidth]{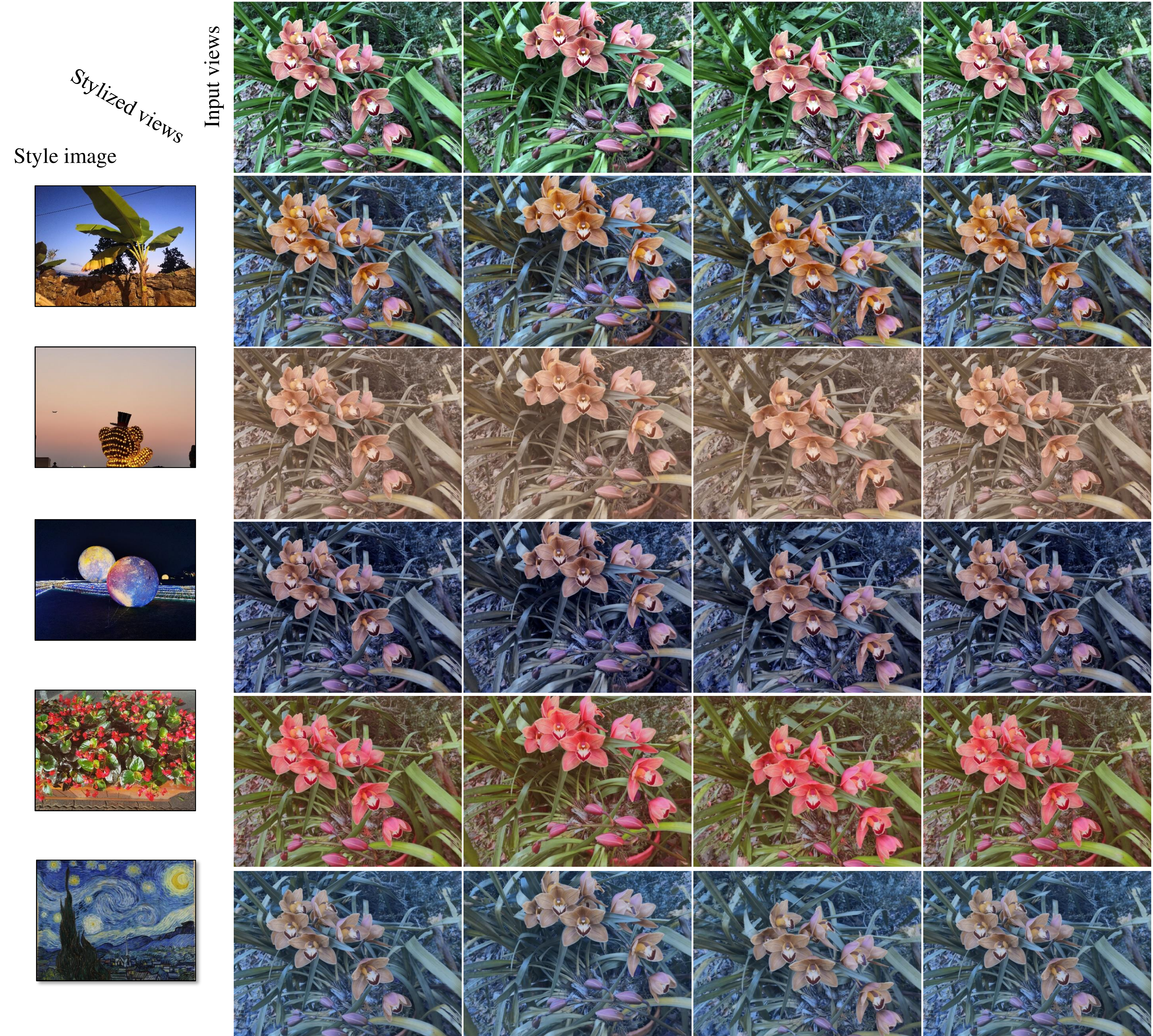}
	\caption{Photorealistic stylization results with the $\operatorname{orchids}$ 3D scene on LLFF dataset.}
	\label{fig:app_s_orchids}
	\vspace{-4mm}
\end{figure*}

\begin{figure*}[htbp]
	\centering
	\includegraphics[width=1.\linewidth]{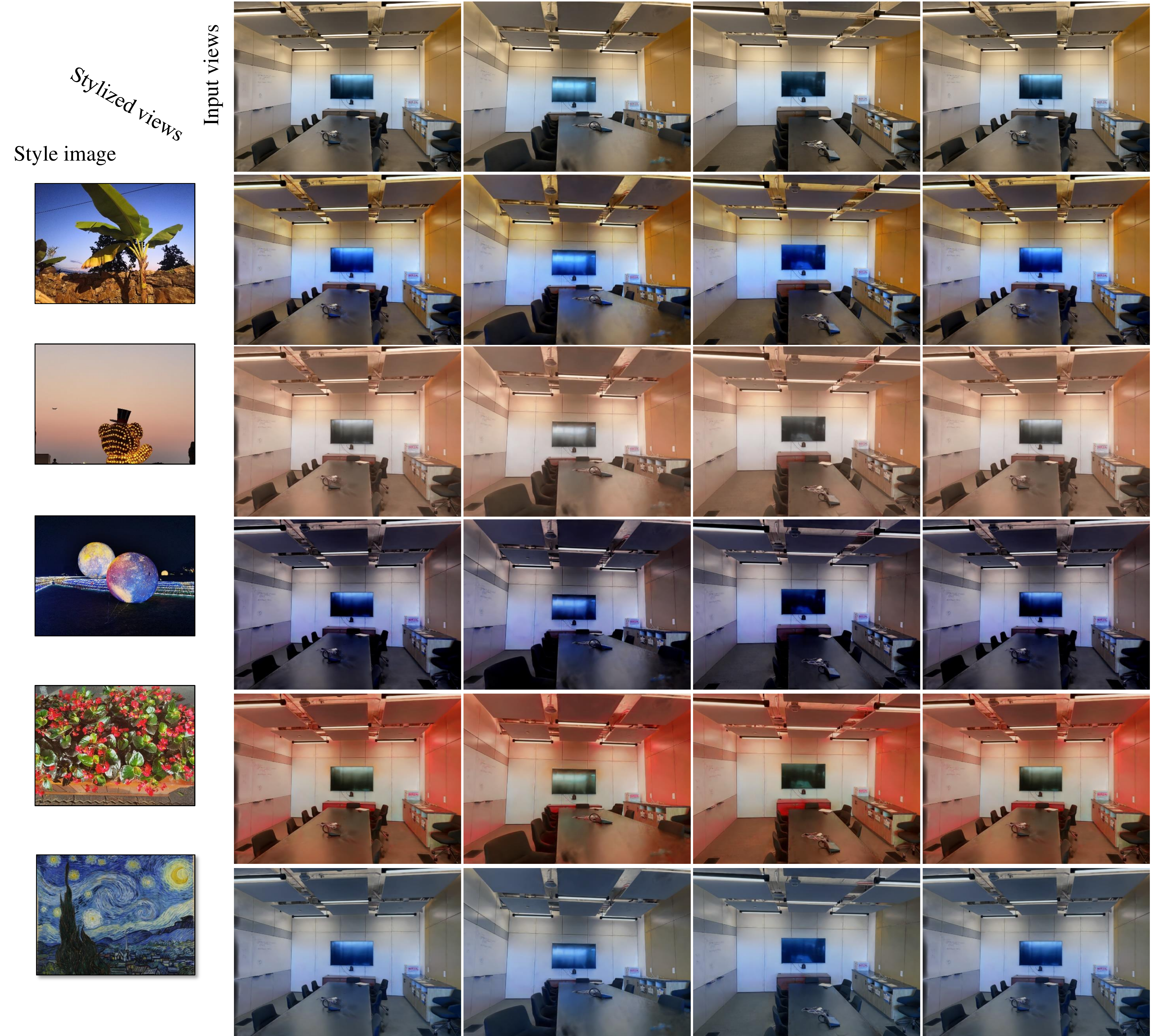}
	\caption{Photorealistic stylization results with the $\operatorname{room}$ 3D scene on LLFF dataset.}
	\label{fig:app_s_room}
	\vspace{-4mm}
\end{figure*}

\begin{figure*}[htbp]
	\centering
	\includegraphics[width=1.\linewidth]{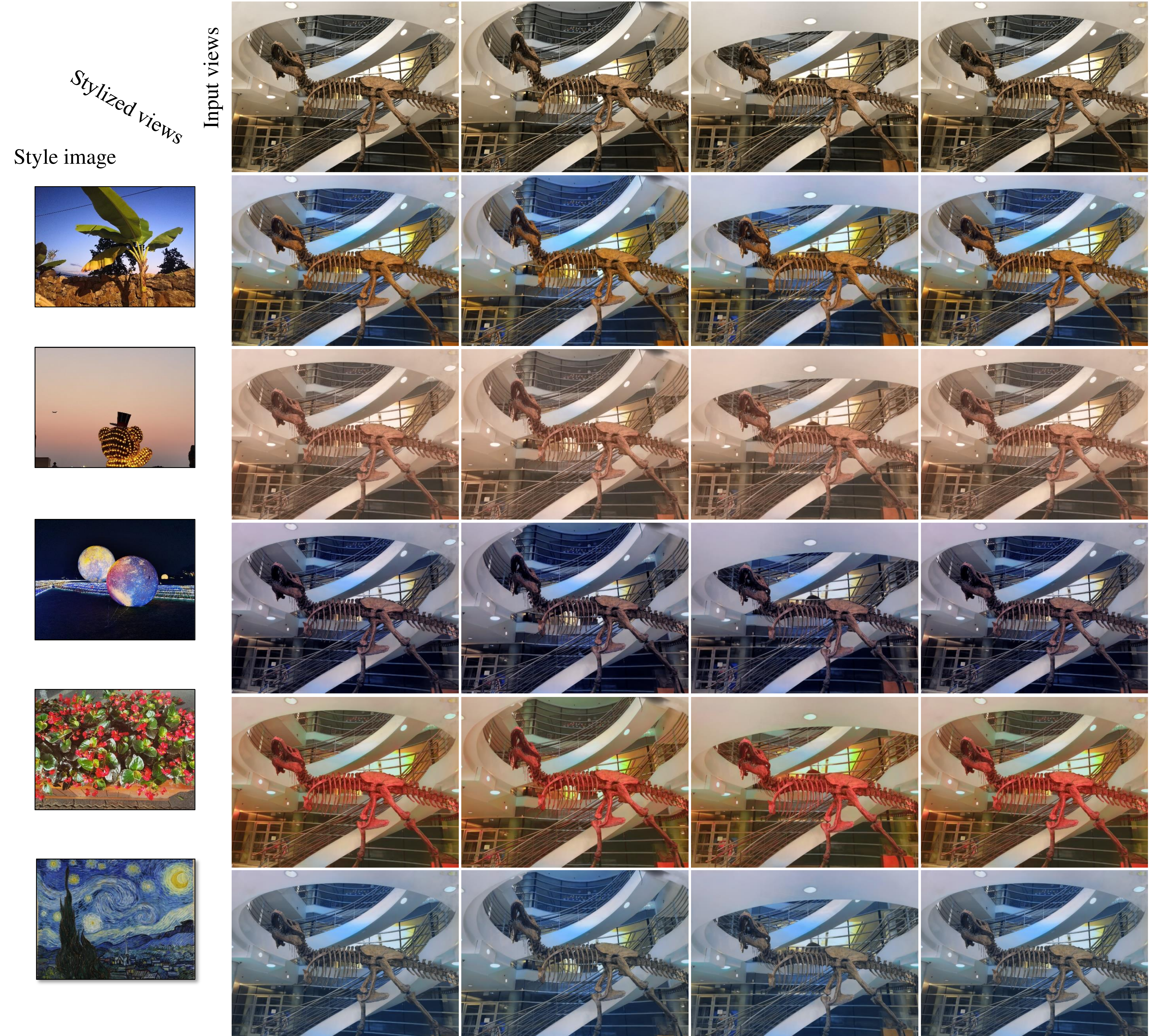}
	\caption{Photorealistic stylization results with the $\operatorname{trex}$ 3D scene on LLFF dataset.}
	\label{fig:app_s_trex}
	\vspace{-4mm}
\end{figure*}

\begin{figure*}[htbp]
	\centering
	\includegraphics[width=1.\linewidth]{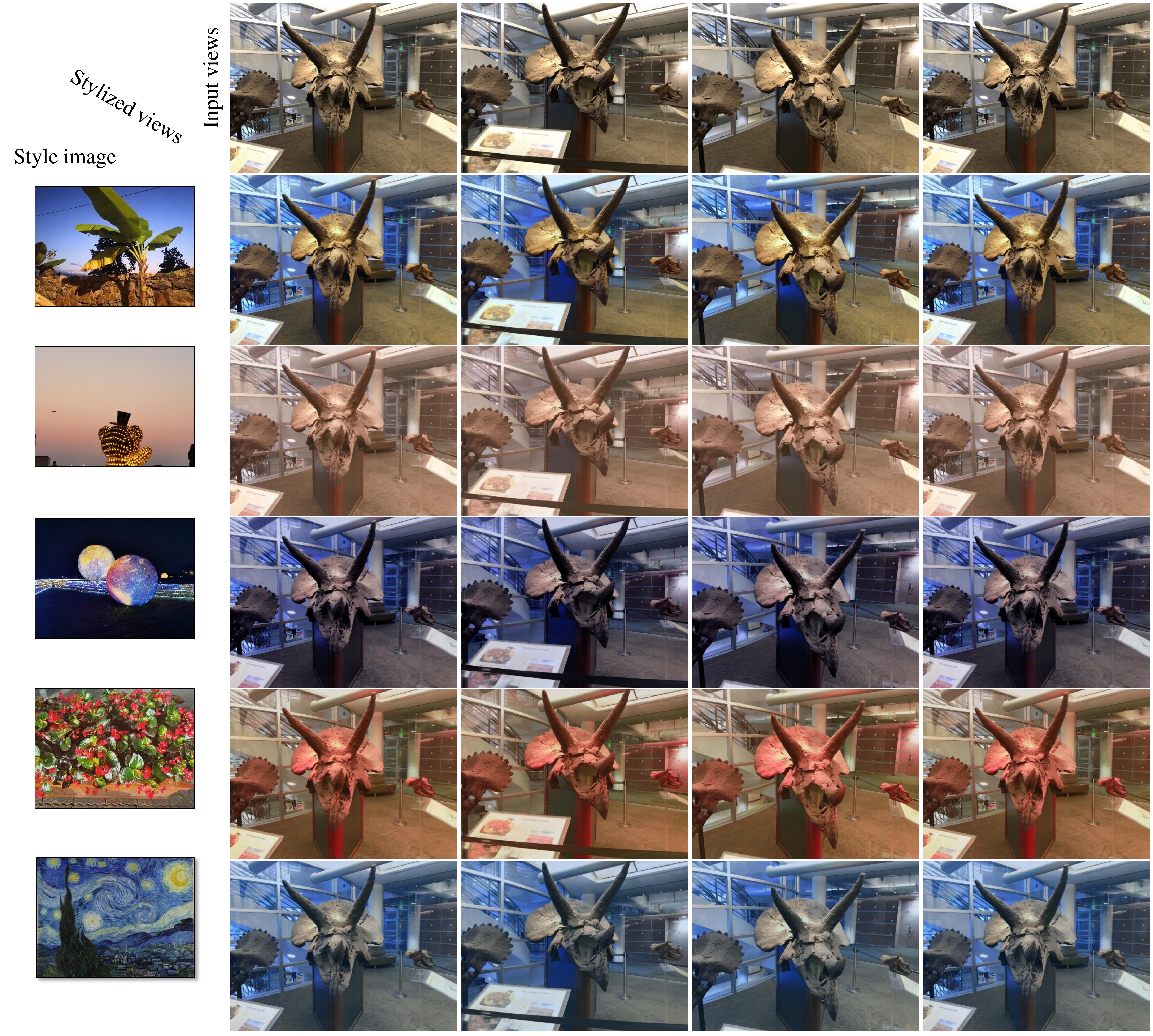}
		\caption{Photorealistic stylization results with the $\operatorname{horns}$ 3D scene on LLFF dataset.}
	\label{fig:app_s_horns}
	\vspace{-4mm}
\end{figure*}

Fig. ~\ref{fig:app_s_fern}, ~\ref{fig:app_s_flower}, ~\ref{fig:app_s_leaves}, ~\ref{fig:app_s_orchids}, ~\ref{fig:app_s_room}, ~\ref{fig:app_s_trex} and ~\ref{fig:app_s_horns} shows more photorealistic stylization results of $\operatorname{fern}$, $\operatorname{flower}$, $\operatorname{leaves}$, $\operatorname{orchids}$, $\operatorname{room}$, $\operatorname{trex}$ and $\operatorname{horns}$ 3D scenes respectively with different style images on Local Light Field Fusion(LLFF) ~\cite{mildenhall2019local} dataset.

\begin{figure*}[htbp]
	\centering
	\includegraphics[width=1.\linewidth]{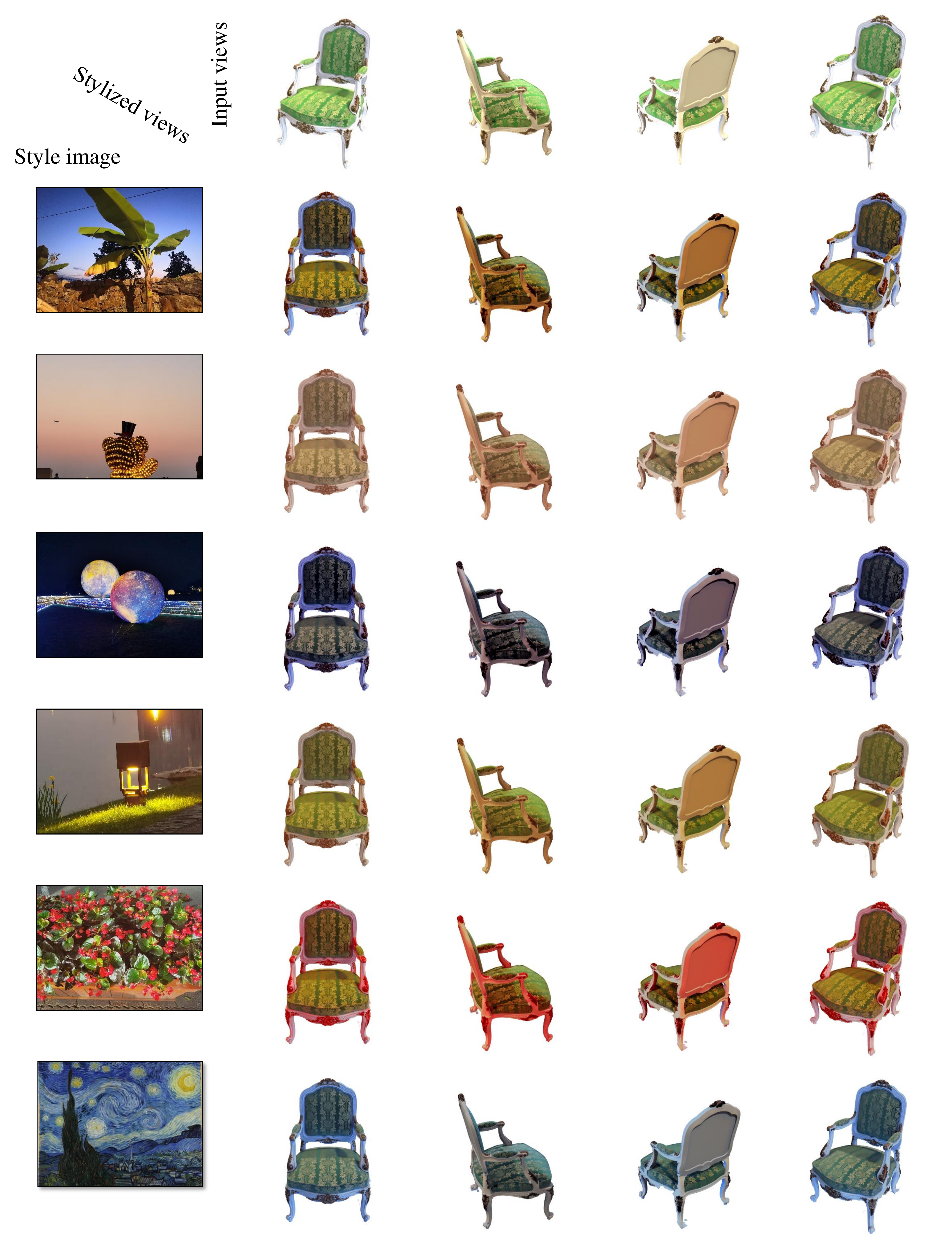}
	\caption{Photorealistic stylization results with the $\operatorname{chair}$ 3D scene on NeRF-Synthetic dataset.}
	\label{fig:app_s_chair}
	\vspace{-4mm}
\end{figure*}

\begin{figure*}[htbp]
	\centering
	\includegraphics[width=1.\linewidth]{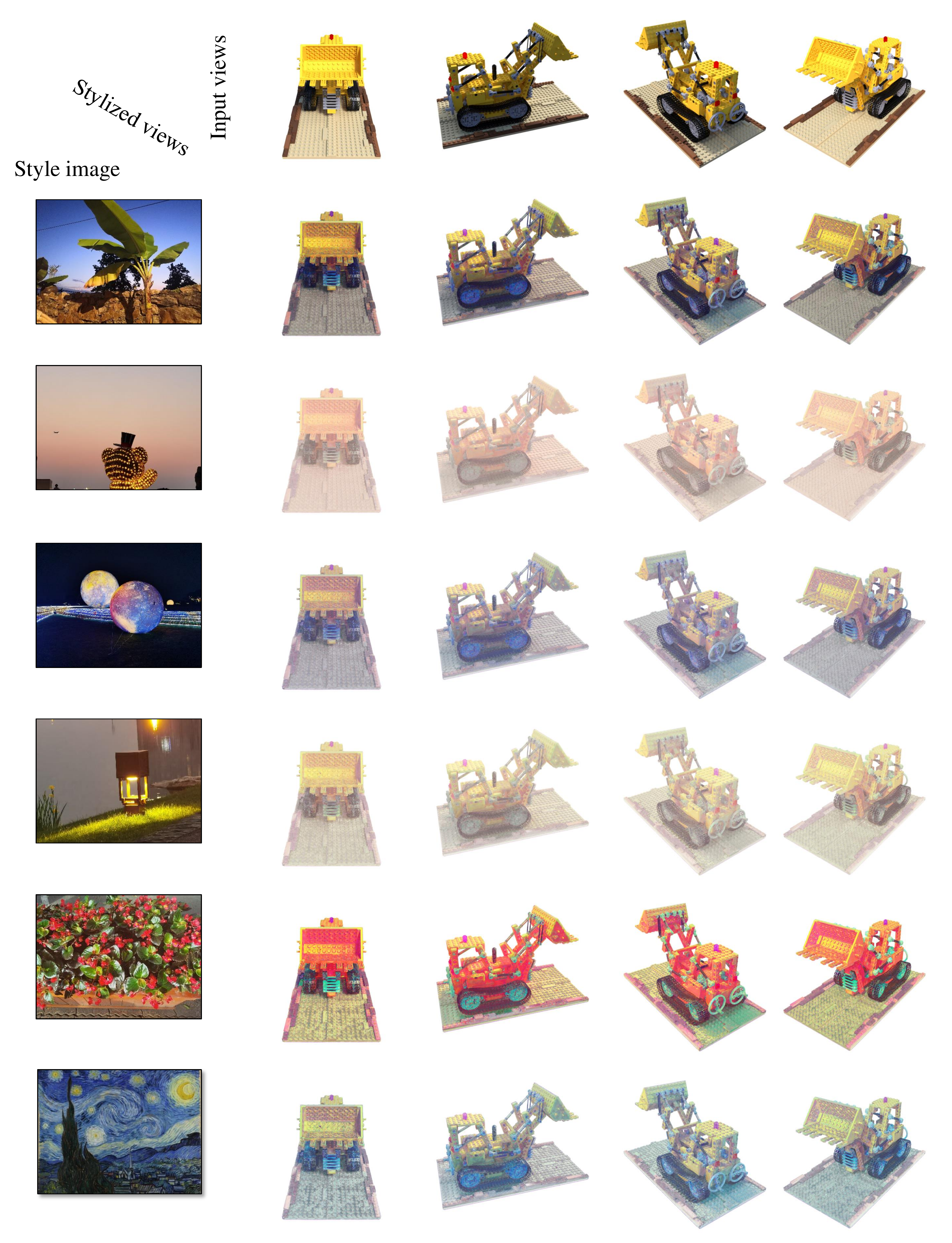}
		\caption{Photorealistic stylization results with the $\operatorname{lego}$ 3D scene on NeRF-Synthetic dataset.}
	\label{fig:app_s_lego}
	\vspace{-4mm}
\end{figure*}

\begin{figure*}[htbp]
	\centering
	\includegraphics[width=1.\linewidth]{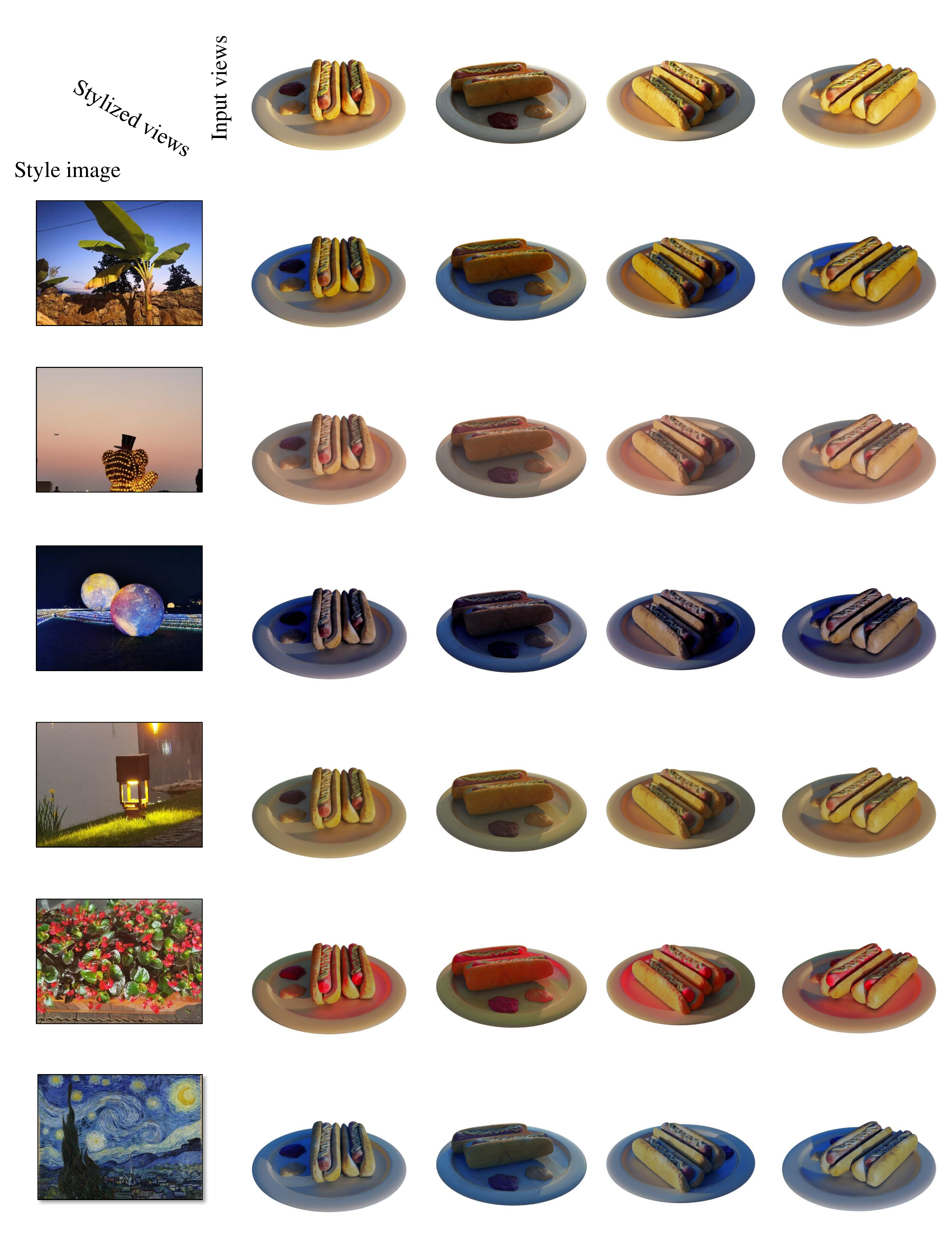}
		\caption{Photorealistic stylization results with the $\operatorname{hotdog}$ 3D scene on NeRF-Synthetic dataset.}
	\label{fig:app_s_hotdog}
	\vspace{-4mm}
\end{figure*}

\begin{figure*}[htbp]
	\centering
	\includegraphics[width=1.\linewidth]{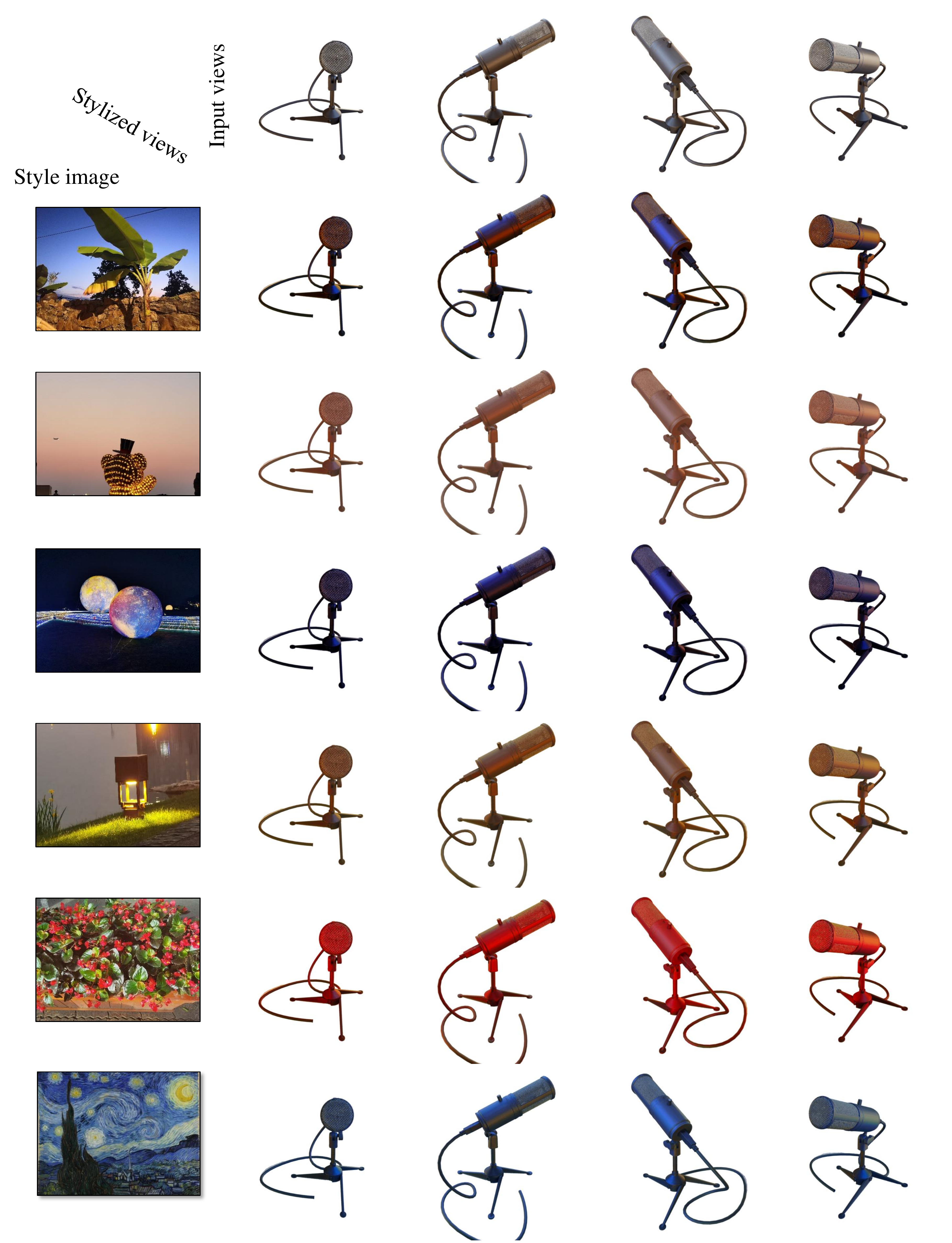}
		\caption{Photorealistic stylization results with the $\operatorname{mic}$ 3D scene on NeRF-Synthetic dataset.}
	\label{fig:app_s_mic}
	\vspace{-4mm}
\end{figure*}

\begin{figure*}[htbp]
	\centering
	\includegraphics[width=1.\linewidth]{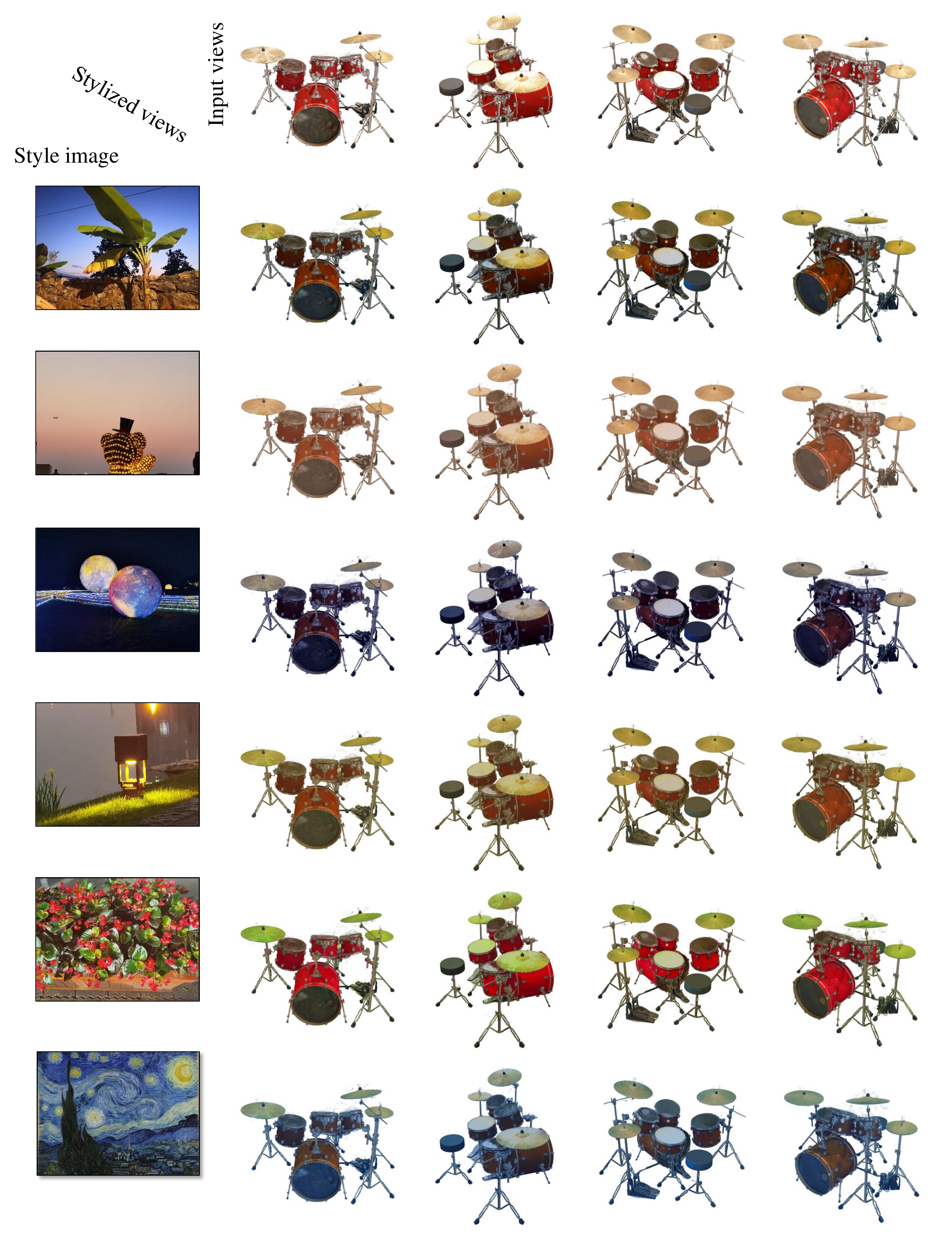}
		\caption{Photorealistic stylization results with the $\operatorname{drums}$ 3D scene on NeRF-Synthetic dataset.}
	\label{fig:app_s_drums}
	\vspace{-4mm}
\end{figure*}

\begin{figure*}[htbp]
	\centering
	\includegraphics[width=1.\linewidth]{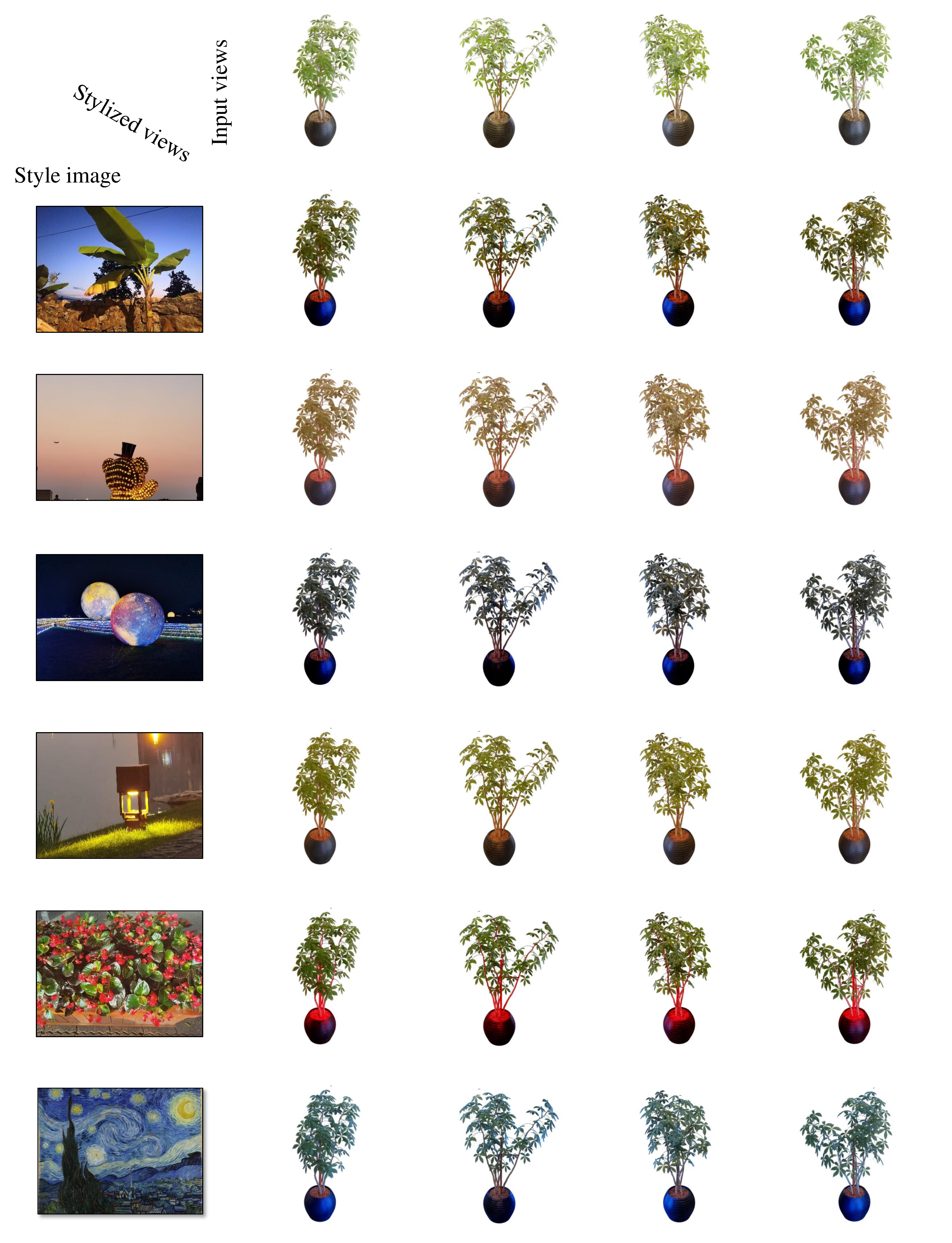}
		\caption{Photorealistic stylization results with the $\operatorname{ficus}$ 3D scene on NeRF-Synthetic dataset.}
	\label{fig:app_s_ficus}
	\vspace{-4mm}
\end{figure*}

Fig. ~\ref{fig:app_s_chair}, ~\ref{fig:app_s_lego}, ~\ref{fig:app_s_hotdog}, ~\ref{fig:app_s_mic}, ~\ref{fig:app_s_drums} and ~\ref{fig:app_s_ficus} shows more photorealistic stylization results of $\operatorname{chair}$, $\operatorname{lego}$, $\operatorname{hotdog}$, $\operatorname{mic}$, $\operatorname{drums}$ and $\operatorname{ficus}$ 3D scenes respectively with different style images on NeRF-Synthetic ~\cite{mildenhall2020nerf} dataset.

From these results, we can see that the color features of different style images will change the color of the 3D scene, which realizes the photorealistic style transfer of the 3D scene and ensures consistency in space.



\end{document}